\documentclass[Afour,sageh,times]{sagej}

\usepackage{moreverb,url}

\usepackage[colorlinks,bookmarksopen,bookmarksnumbered,citecolor=blue,urlcolor=blue]{hyperref}

\usepackage{xcolor}
\usepackage{graphicx}
\usepackage{amsmath}
\usepackage{amssymb}
\usepackage{multirow}
\usepackage{booktabs}
\usepackage{algorithm}
\usepackage{algpseudocode}
\usepackage{gensymb}
\usepackage{rotating}
\usepackage{subcaption}
\DeclareMathOperator*{\argmax}{argmax}

\newcommand\BibTeX{{\rmfamily B\kern-.05em \textsc{i\kern-.025em b}\kern-.08em
T\kern-.1667em\lower.7ex\hbox{E}\kern-.125emX}}

\def\volumeyear{2016}

\begin{document}

\runninghead{Huang et al.}

\title{Why semantics matters: A deep study\\ on semantic particle-filtering localization in a LiDAR semantic pole-map}

\author{
Yuming Huang\affilnum{1}, Yi Gu\affilnum{1}, Chengzhong Xu\affilnum{1} and Hui Kong\affilnum{2}
}

\affiliation{\affilnum{1}The State Key Laboratory of Internet of Things for Smart City (SKL-IOTSC), Department of Computer Science, University of Macau, Macau\\
\affilnum{2}The State Key Laboratory of Internet of Things for Smart City (SKL-IOTSC), Department of Electromechanical Engineering, University of Macau, Macau
}

\corrauth{Hui Kong, 
Faculty of Science and Technology, University of Macau, Macau.}

\email{huikong@um.edu.mo}

\begin{abstract}
   In most urban and suburban areas, pole-like structures such as tree trunks or utility poles are ubiquitous. These structural landmarks are very useful for the localization of autonomous vehicles given their geometrical locations in maps and measurements from sensors.  In this work, we aim at creating an accurate map for autonomous vehicles or robots with pole-like structures as the dominant localization landmarks, hence called pole-map. In contrast to the previous pole-based mapping or localization methods, we exploit the semantics of pole-like structures. Specifically, semantic segmentation is achieved by a new mask-range transformer network in a mask-classfication paradigm. With the semantics extracted for the pole-like structures in each frame, a multi-layer semantic pole-map is created by aggregating the detected pole-like structures from all frames. Given the semantic pole-map, we propose a semantic particle-filtering localization scheme for vehicle localization. Theoretically, we have analyzed why the semantic information can benefit the particle-filter localization, and empirically it is validated on the public SemanticKITTI dataset that the particle-filtering localization with semantics achieves much better performance than the counterpart without semantics when each particle's odometry prediction and/or the online observation is subject to uncertainties at significant levels.
\end{abstract}


\maketitle

\section{Introduction}
\label{sec:intro}


The precise acquisition of vehicle pose in realtime in an urban or suburban environment 
enables autonomous vehicles to plan a path and navigate to a specified destination location.
While GPS can provide accurate position estimates at a
global scale, it does not provide sufficiently
accurate estimates because it suffers from substantial errors due to multi-path
effects in urban canyons or signal degradation caused by occlusions by trees. 
A popular alternative way to autonomous vehicle
localization is based on matching sensor observations against the data already saved in a previously acquired map.

In many existing methods, the same type of sensor is usually used during mapping creation and during vehicle localization given the map. 
Cameras are generally cheap and lightweight sensors and are available widely.  The monocular camera is not capable of providing absolute range information. Therefore, during the mapping process, binocular cameras can be adopted \citep{orbslam3} or a monocular camera can be combined with a multiple-channel LiDAR sensor \citep{LightWeightSemantic, VisualLoc-in-Lidar2} if camera sensors are preferably selected for localization. Generally, localization based on cameras is sensitive to illumination variations and view angles, although relying on matching the geometry of a sparse set of 3D points reconstructed from image features with the map's point cloud \citep{VisualLoc-in-Lidar2} or seeking illumination-invariant image features for matching sequences \citep{life-long-ICRA15}.

In contrast, LiDAR sensors are almost insensitive to external light conditions and can provide accurately
range measurements. 
Our method exploits using a multiple-channel LiDAR sensor during both 
mapping creation and during localization given the map. This type of sensor setting for both mapping and localization has been widely used so far \citep{LOAM-RSS15, Lego-LOAM, overlapNet}. In general, the existing LiDAR-based map representation has a large memory cost although some existing approaches can achieve a significant memory consumption by only saving extracted high-curvature/corner points in the map. 

Our method proposed in this paper belongs to the route of LiDAR-based mapping and localization approaches. 
Specifically,  we aim at creating an accurate map for autonomous vehicles or robots with pole-like structures as the dominant localization landmarks, hence called pole-map. In contrast to the previous pole-based  maps or ego-motion estimation methods, we exploit the semantics of pole-like structures. Specifically, semantic segmentation can be achieved by a new mask-range network, where pole-like structures with semantics are obtained in each frame. The semantics are utilized in the offline mapping process. Given the semantic pole-map, we propose a semantic particle-filtering localization method for vehicle localization. It is shown that when the uncertainty of each particle's prediction is subject to a nonlinear increase or abrupt change, the particle-filtering localization with semantics achieves much better performance than the counterpart without semantics.

Our contributions are threefold:
\begin{itemize}
  \item We propose a relatively complete framework for semantic mapping and localization where the localization is achieved based on semantic particle filtering in a semantic pole-map created offline by a multi-channel LiDAR sensor. 
  
  \item For the offline semantic pole-map creation, based on LiDAR's range-view representation, we first propose a semantic segmentation transformer network in a mask-classification paradigm to segment pole-like structures from LiDAR scans. Then a multi-layer semantic pole-map is created by aggregating the detected pole-like structures based on the vehicle's ego-motion with semantic-feature embeddings. 
  
  \item For the online vehicle localization given the created semantic pole-map, we theoretically analyze how the semantic information can benefit the particle-filter localization. Different from existing works, our method utilizes both the geometric and semantic discrepancy to improve particle-filter localization by both improving its proximity to the truth pose and discouraging the bad proposal of poses. Empirically, we have demonstrated its effectiveness and improvement over conventional methods based on the multi-layer semantic map in the real-world SemanticKITTI dataset with simulated uncertainty. 
\end{itemize}

\section{Related Works}  
\label{sec:related}

Different types of sensors have been utilized for vehicle localization given a 3D map. These sensors can be used individually, although in many applications they compensate each other as
well as with other sensors (e.g. Inertial Measurement Unit
(IMU), wheel odometry, radar) for optimal localization performance. However, for true redundancy, these
sub-systems must work independently as well to avoid
dangerous situations from sensor malfunction. In this paper, we are only interested in vehicle localization in cities or suburbs.  Specifically, we only review localization in a prior point-cloud map 
with an individual camera (including a stereo camera) or a multiple-channel LiDAR sensor. 
The related works on indoor localization are beyond the scope of this work. 

\textbf{Camera-based localization within a 3D map}: Generally, given a 3D map, camera-based localization is cheaper than LiDAR-based localization. Compared with the LiDAR-based method, monocular or RGB-D cameras are generally
sensitive to illumination variation, and seasonal and adversarial weather changes. 

\cite{VisualLoc-in-Lidar1} proposed a camera-based localization within a 3D prior map
(augmented with surface reflectivities) by a 3D LIDAR sensor. Given an initial pose, quite a few
synthetic views of the environment are generated from the 3D prior map, camera localization is achieved by matching the live camera view with these synthetic views based on normalized mutual information. 
\cite{VisualLoc-in-Lidar1a} propose an online 6-DoF
visual localization across a wide range of outdoor illumination
conditions throughout the day and night using a 3D scene prior. An illumination-invariant image representation is adopted with a normalized information distance as the matching measure between the online image and the colored 3D LiDAR point cloud. 

Another way to deal with illumination variation is based on matching point clouds. \cite{VisualLoc-in-Lidar2} propose to locate a monocular camera in a prebuilt point cloud map by a LiDAR sensor. They reconstruct a sparse set of 3D points from image features based on local bundle adjustment, which are continuously matched against the map to track the camera pose in an online fashion. 
Similarly, \cite{VisualLoc-in-Lidar3} also propose to continuously estimate similarity transformations that align the 3D structure reconstructed by visual SLAM to the point cloud map. 
Stereo cameras have also been applied for vehicle localization. \cite{VisualLoc-in-Lidar4} aim to achieve the consumer level
global positioning system (GPS) accuracy by
matching the depth from the stereo disparity with 3D LiDAR maps. 

Aside from pure geometric clues, semantic information has also been exploited extensively for both efficiency and accuracy of localization. 
\cite{Laneloc} propose to use lanes as localization cues. Toward this goal, they manually
annotated lane markers in the LiDAR intensity map and these lane markers are then detected online using a stereo camera, and matched against the ones on the map.
\cite{trafficsign1} and \cite{trafficsign2} utilize traffic signs to implement image-based localization.
\cite{Semantic1} built dense
semantic maps using image segmentation and conducted
localization by matching both semantic and geometric cues.
To save map size, \cite{sparsesemantic} formulate localization problem in a Bayesian filtering framework, and exploit
lanes, traffic signs, as well as vehicle dynamics to localize robustly with respect to a sparse semantic map. A similar work is proposed in \cite{LightWeightSemantic} which exploits semantic road elements for lightweight map-based camera localization.

Besides corner points, lines have also been used for localization. In \cite{2D-3Dlines}, correspondence between 2D lines in video sequences and 3D lines are established for a rough 
Monocular Camera Localization in Prior LiDAR Maps with 2D-3D
Line Correspondences. 
\cite{2D-3Dlines2} also propose a vehicle relocalization method in a 3D line-feature map using Perspective-n-Line given a known vertical direction. 

In addition, integration of more than one type of sensor can also improve localization accuracy, e.g., radar plus LiDAR in fire-disaster scenes \citep{Radar-Lidar-in-LiDARprior}.
\cite{sensorfusion} propose a sensor-fusion-based localization method in a 3D map, where they adaptively use information
from complementary sensors such as GNSS, LiDAR, and
IMU to achieve high localization accuracy and resilience in
challenging scenes, such as urban downtown, highways, and
tunnels.

\textbf{LiDAR-based localization within a 3D map}:
Due to the fact that LiDAR sensors can provide high-precision data irrespective of the distance measured, LiDAR sensors have also been widely adopted in localization within a given 3D map. 
\cite{scan-feature} propose a feature quantity for scan
data based on the distribution of clusters for localization based on the lidar data and a precise 3-D map. 
\cite{JFR10} propose the Multi-frame Odometry compensated
Global Alignment (MOGA) algorithm to globally localize a rover by matching features from a three-dimensional (3D) orbital elevation map to features from the rover-based 3D LIDAR scans.

\cite{TITS15} propose to extract road-marker and curb features from multiple-channel LiDAR data and these features are
stored in a high-resolution occupancy grid map and a Monte-Carlo localization (MCL) is used for vehicle localization. Similarly, \cite{CurbICRA12} introduce an MCL method using the curb-intersection features on urban roads with a single tilted 2D LIDAR. 
\cite{double-layer} also propose using curb and vertical features for localization. 
\cite{overlapNet} propose a neural network-based 
observation model that predicts the overlap and yaw
angle offset between the online LiDAR reading and virtual
frames generated from a pre-built map. 

Besides, LiDAR data have also been converted into an image-like representation and used to match the online data with the ones stored in the map for vehicle localization, e.g., Scan Context \citep{scancontext}, LiDAR Iris \citep{lidariris}, Intensity Scan Context \citep{intensityscancontext}. These methods can be used for rough localization instead of at centimeter-level accuracy.


\subsection{Pole-based Mapping and Localization}
Pole-like structures have also been used as landmarks in mapping and localization because of their invariance over time and across viewpoints. \cite{spangenberg2016pole} propose to extract poles from depth images by a stereo camera. However, depth estimation from stereo cameras is sensitive to illumination conditions. \cite{sefati2017improving} use a LiDAR in addition to a stereo camera to extract the poles. With the popularization of LiDAR, recently, some methods have been proposed to only utilize LiDAR for localization. \cite{weng2018pole,li2021robust,lu2020pole,chen2021pole} propose to extract poles from voxelized point clouds. \cite{chen2021pole} incorporate the curb extracted from the Birds-Eye View (BEV) projection to achieve pole-curb fusion for localization. \cite{schaefer2019long,schaefer2021long} explicitly model the occupied and free space by ray tracing that considers both the start point and end point. However, their runtime is limited as the 3D voxel is computationally consuming. In contrast, \cite{dong2021online,dong2023online} use range-view-based methods to extract the poles. They use a series of geometric heuristics to cluster the poles in range view and fit a circle for each pole to obtain the center position of the pole on the global map. \cite{wang2021pole} train the RangeNet++ \citep{milioto2019rangenet++} for pole segmentation on they own labeled dataset. \cite{dong2023online} propose to train the SalsaNext \citep{cortinhal2020salsanext} for pole segmentation using the pole labels generated by the geometric heuristics across several datasets and achieve higher localization performance than directly using the generated labels. 
Different from those works, we use the mask-classification paradigm to segment the poles to achieve better mapping performance without introducing many computational costs. 

Having a map built from the extracted poles, these works use non-linear optimization or a Monte Carlo particle filter in localization. \cite{schaefer2019long,schaefer2021long,dong2021online,dong2023online} use the Monte Carlo particle filter to estimate the pose with the nearest neighbor searching as correspondence between observation and landmarks in the map. \cite{li2021robust} use 4D vector including position, radius, and height for corresponding pole finding. \cite{chen2021pole} propose a Branch-and-Bound-based global optimization method to tackle the data association problem of poles and use a non-linear optimization method to fuse the pole cost and curb cost to obtain the vehicle location. 
\cite{wang2021pole} propose to match the poles between local and global maps according to the semantic and geometric consistency of the poles. After finding the correspondences, the Iterative Closest Point \citep{besl1992method} algorithm on the pole centroid and pole point cloud is utilized to optimize the pose for relocalization, and then combined with the LiDAR odometry for localization. 
In contrast, we incorporate semantic information predicted from the segmentation network into the Monte Carlo particle filter for localization.

\subsection{Particle-Filter Localization with Semantics}

The distance field \citep{jiang2021particle,miller2021any,akai2020semantic} is utilized to describe the nearest distance from obstacles or surfaces in each semantic category.  However, the map representation is point-wise and might be memory inefficient.
Observed objects match the ones in the built map with the corresponding semantic category \citep{bavle2018stereo,zimmerman2022long}, where they depend on camera-based detection and recognition in indoor environments. All these methods utilize the geometric discrepancy (position and orientation difference) and incorporate semantic information into the association for better geometric discrepancy calculation.

In \cite{jeong2020hdmi,bernuy2018topological}, the particle weights are updated by calculating the discrepancy in semantics between the online observations and the ones in the built map. 
In \cite{jeong2020hdmi}, the semantic discrepancy is measured by bitwise AND operation of segmented labels in bird-eye-view projection by images. 
In \cite{bernuy2018topological}, the semantic features are represented by histograms of the number of segmented labels in images, and the cosine similarity between these features is used as the discrepancy between online observations and the ones in the built topological map. 
It is noted that these methods adopt cameras as the main sensor and are sensitive to illumination variations.
In \cite{yan2019global}, semantic descriptors are extracted from LiDAR scans as observations and compared with map information from the OpenStreetMap \citep{OpenStreetMap} with Hamming distance as the discrepancy. However, only the semantic discrepancy is utilized in this work and the absence of geometric discrepancy might limit the localization performance.
Different from the existing works, we utilize both the semantic and geometric discrepancy of pole-like landmarks, extracted by the LiDAR sensor, which is efficient and robust to illumination.

\section{Our Method}  
In this section, we introduce our methods for extracting pole-like objects, building pole-maps, and vehicle localization in the created pole-maps. To extract pole-like objects, we first segment the LiDAR scans into discrete regions to distinguish pole-like objects and other objects and then cluster the segmented pole-like objects. While building the pole-map, we convert the extracted pole-like objects into map landmarks with geometric and semantic information. In localization, we propose the semantic-aware Monte Carlo particle filter to improve accuracy and robustness. We have shown that when pole-like objects are very sparsely distributed in the map or when the prediction of each particle's pose is subject to large uncertainty, semantics play a key role in improving localization accuracy.

\begin{figure*}[htbp]
    \centering
	\includegraphics[width=0.99\textwidth]{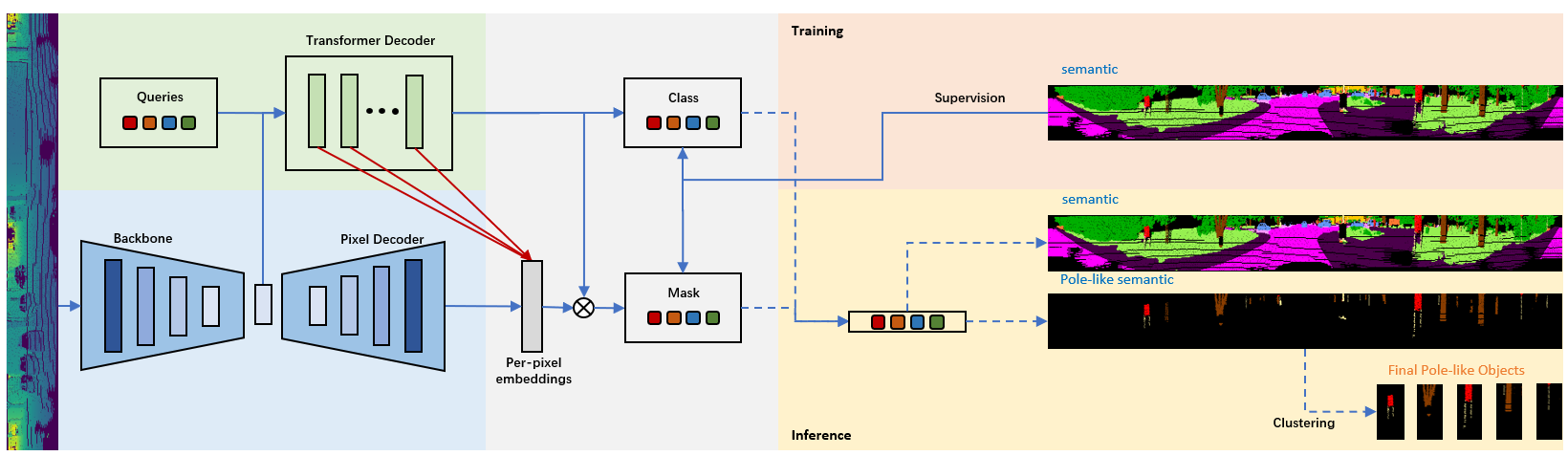} 
	\caption{The architecture of pole segmentation network. The left is the network design. The top right is the training procedure. The bottom right is the inference stage.}
	\label{network}   
\end{figure*}

\subsection{Pole Segmentation}
Deep neural networks are popular in semantic prediction. To obtain the pole-like objects with semantics from LiDAR scans, one can use an object detection network to predict the semantic category with a bounding box for each pole-like object, or a segmentation network to predict the semantic category for each point. The former directly predicts the semantics with geometric information, and the latter provides the semantic information for each point for further processing. We choose the latter because it can estimate more accurate pole-like instances by geometric clustering with per-point semantics than directly using the bounding box, as the bounding box is a cube rather than a cylinder and it is difficult to generalize it to predict the parameters of an object's shape.


An efficient representation of the LiDAR point cloud for LiDAR-based segmentation is the range-view image. By this representation, the LiDAR point cloud is projected into a range-view image according to spherical projection as follows,
\begin{equation}
\left(
\begin{array}{ccc}
    u\\
    v\\
    \end{array}
\right) = 
\left(
\begin{array}{ccc}
    \frac{1}{2}\left[1-arctan(y, x)\pi^{-1}\right]W\\
    \left[1-(arcsin(zr^{-1})+f_{up})\frac{1}{f}\right]H\\
\end{array}
\right),
\end{equation}
where W and H are the width and height of the range image, respectively. $f = f_{up} + f_{down}$ is the LiDAR’s vertical field-of-view. The range value $r=\sqrt{x^2+y^2+z^2}$ is calculated according to the point coordinates $\left[x, y, z\right]^T$ and $(u, v)^T$ are the image coordinates in the range view. The range-view image representation is used to predict the pixel-wise category, which is then projected to the original point cloud to obtain the point-wise category. We chose the range-view-based representation because it is generally more efficient compared with point-based and voxel-based semantic segmentation methods.

\subsubsection{LiDAR Segmentation by Mask-Classification}
\hfill

Following \cite{long2015fully}, previous range-view-based methods \citep{milioto2019rangenet++,cortinhal2020salsanext,zhao2021fidnet} mostly predict the pixel-wise category in the range-view image by per-pixel classification. Differently, we use the mask-classification paradigm \citep{cheng2021per} to predict the region for each category as well as the pixel-wise category. Mask classification achieves state-of-the-art performance in image segmentation \citep{cheng2021mask2former}, however, few works have investigated its effectiveness in LiDAR-based segmentation. 
 In this work, we investigate the mask-classification paradigm in the semantic segmentation of pole-like objects for semantic mapping. 
Specifically, we use SalsaNext \citep{cortinhal2020salsanext} network structure as the backbone and the same transformer decoder as in \cite{cheng2021per}. 
In the training procedure, the supervision loss combines the classification loss $L_{cls}$ and binary mask loss $L_{mask}$ as in \cite{cheng2021per},
\begin{equation}
\label{eq:maskcls}
L_{total} = L_{cls}+L_{mask}.
\end{equation}
In \cite{cheng2021per}, the classification loss $L_{cls}$ is the cross-entropy loss and the binary mask loss $L_{mask}$ combines the focal loss \citep{lin2017focal} and the dice loss \citep{milletari2016v}. In addition to them, we add Lov\'asz loss \citep{berman2018lovasz} to $L_{mask}$ to directly optimize the Jaccard index, which is shown effective in LiDAR data in \cite{cortinhal2020salsanext}, and remove the supervision of pixels invalid in the range-view image from $L_{mask}$.
For the inference procedure, the class prediction and mask prediction are combined to obtain the final category for each pixel in the same way as \cite{cheng2021per}.
The network is trained by all semantic categories from labels, and we focus on the performance of pole-like categories, e.g., pole, trunk, and traffic sign.
The network architecture is shown in Fig. \ref{network}. Using the mask-classification paradigm with data augmentation which is introduced in the following section, it has been shown in the experimental section that the performance of pole-like object segmentation is superior to the previous range-view based methods \citep{cortinhal2020salsanext,zhao2021fidnet}.

\subsubsection{Data Augmentation for Mask-Classification} 
\hfill

In the training procedure of LiDAR-based segmentation, simple data augmentations for LiDAR point cloud have been widely used, including random rotation, translation, flipping, and point dropping. However, with these common augmentations, the performance of mask classification cannot be on par with that of its per-pixel baseline counterpart \citep{cortinhal2020salsanext}. 
Although the mask-classification paradigm achieves unified semantic and panoptic segmentation with SOTA performance in RGB image \citep{cheng2021per,cheng2021mask2former}, the mask-classification paradigm still has the following issues when applied to LiDAR's range-view based segmentation. 

\begin{enumerate}
\item The mask-classification paradigm demands a large amount of diverse training data. The publicly available LiDAR segmentation datasets are relatively limited and small, thus cannot satisfy the requirement of transformer architectures \citep{vaswani2017attention, DBLP:conf/iclr/DosovitskiyB0WZ21} which usually require more training data compared with the convolutional neural networks. 
\item The mask-classification paradigm relies heavily on contextual cues. Limited datasets are often biased and mislead neural networks to find shortcuts during training and result in poor model generalization ability to rare or unseen situations \citep{nekrasov2021mix3d, DBLP:conf/cvpr/ShettySF19}, especially for mask-classification paradigm since the transformer decoder only queries the global context embeddings to find objects. Unfortunately, pole-like things are often related to context knowledge, e.g., poles are often present at the side rather than in the middle of the road.
\item Very biased to the high-frequency classes. Mask-classification paradigm makes predictions with queries whose number is larger than the categories. The supervision of this procedure is query-wise rather than pixel-wise, thus a mistake in the category of a query usually results in mistakes in a large area. Because the LiDAR data collected in driving scenarios are very long-tailed, e.g., points in pole-like categories are evidently less than some other dominant categories such as road, the network may perform poorly in the rare classes.
\end{enumerate}

\begin{figure}
     \centering
     \begin{subfigure}[b]{0.49\columnwidth}
         \centering
         \includegraphics[width=\columnwidth]{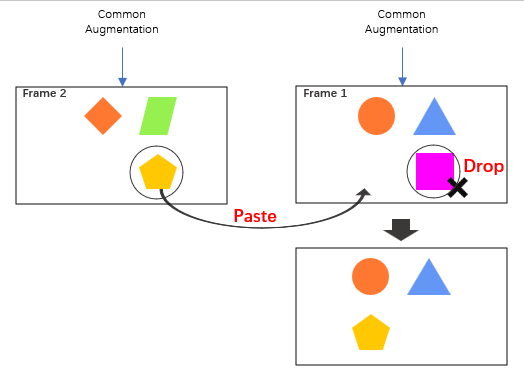}
         \caption{}
         \label{fig:wpd1}
     \end{subfigure}
     \hfill
     \begin{subfigure}[b]{0.49\columnwidth}
         \centering
         \includegraphics[width=\columnwidth]{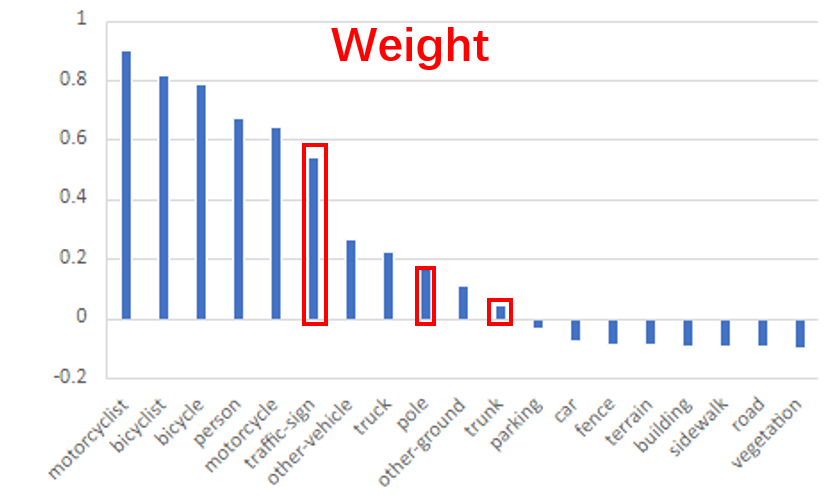}
         \caption{}
         \label{fig:wpd2}
     \end{subfigure}
     \hfill
     \begin{subfigure}[b]{0.98\columnwidth}
         \centering
         \includegraphics[width=\columnwidth]{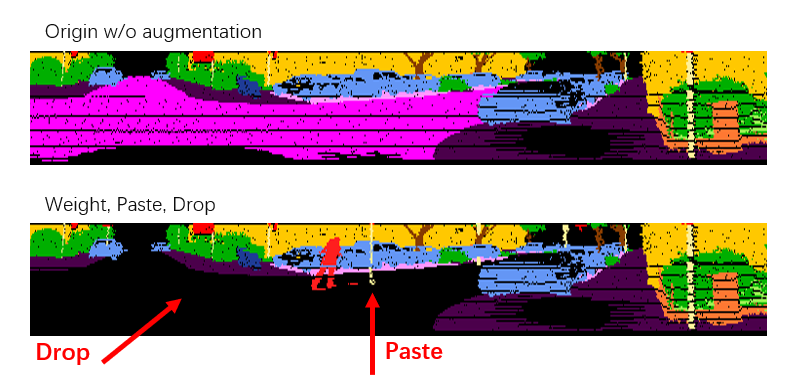}
         \caption{}
         \label{fig:wpd3}
     \end{subfigure}
\caption{(a) Paste and Drop Operation. (b) Weight for Paste and Drop Operation. (c) An example of our Weighted Paste and Drop Data Augmentation. In this example, the pole is pasted into the middle of the road and the road is dropped.}
        \label{fig:wpd}
\end{figure}

To deal with the aforementioned problems, we propose a novel data-augmentation method, which is constituted of three meta-operations: ``Weighted'', ``Paste'' and ``Drop''. The procedure is shown in \ref{fig:wpd}. Initially, two frames are randomly selected from the dataset (noted as the first and second frames), and the common data augmentation is applied. The \textbf{paste} operation first selects the long-tailed objects from the second frame, then adds them to the first frame. The \textbf{drop} operation selects the non-long-tailed class points in the first frame, then deletes these points \ref{fig:wpd1}. The \textbf{weighted} operation is to add a probability to the \textbf{paste} and \textbf{drop} \ref{fig:wpd2}.  

Our Weighted Paste Drop (WPD) data augmentation significantly enlarges the size and diversity of the dataset. To prevent our model from relying on too many contextual cues, our WPD scheme weakens the role of the context priors, as shown in \ref{fig:wpd3}. The ``paste" operation can create unusual or even impossible scene scenarios for the training set. In this example, a ``pole" is ``pasted"  to the middle of a road, which never appears in the original dataset and cannot be created by standard data augmentation. By this way, the ``paste" operation can weaken the context priors. Further, the ``drop" process directly reduces the context bias by removing the area with high context information. In this example, the network is trained to recognize a ``pole" without a sidewalk in its vicinity. To make the dataset less biased, we drop the high-frequency classes, such as road and car, with high probability and paste less-frequency classes for the pole-like objects frequently. In the experiments, we show that the WPD data augmentation improves the performance of both the baseline method and ours. In contrast, the improvement of our method is more evident, indicating that the mask classification is effective in semantic segmentation for our pole-like objects with our data augmentation.

\subsection{Extraction of Pole Information}
We choose $K$ pole-like categories from semantic segmentation as the classes for pole-like objects, e.g., poles, trunks, and traffic signs. After training, the segmentation model predicts the region for pole-like objects with the corresponding category labels. We use DBSCAN \citep{ester1996density} to cluster the pole-like objects from the region corresponding to these categories. Then we extract the geometric and semantic information for the clustered objects. 

Let $\mathbf{P} = \{\mathbf{p}^1,..., \mathbf{p}^N\}$ represent all the $N$ points in a scan, $\mathbf{P_i} = \{\mathbf{p}_i^1,..., \mathbf{p}_i^n\}$ represent the $n$ points belonging to the $i^{th}$ pole-like object, each element of $\mathbf{C_i}$ = $\{\mathbf{c}_i^1,...,\mathbf{c}_i^n\}$ represent the predicted probability vector of each point belonging to the $K$ classes in $\mathbf{P_i}$. 

To estimate the pole geometry, we follow \cite{dong2021online,dong2023online} to use the least-squares circle fitting \citep{bullock2006least} to obtain the position $l = (l_x, l_y)$ and radius $r$ for the $i^{th}$ pole-like object. First, $\mathbf{P_i}$ is projected onto the $XY$ (horizontal) plane to obtain the $X$ coordinates $U_i = \{ u_i^1, ..., u_i^n \}$ and $Y$ coordinates $V_i = \{ v_i^1, ..., v_i^n \}$. Then the relative coordinates $(u_c, v_c)$ of circle center is obtained by solving the following equation:
\begin{equation}
\begin{aligned}
\label{circle_center0}
S_{uu}u_c + S_{uv}v_c & = (S_{uuu} + S_{uvv}) / 2,\\
S_{uv}u_c + S_{vv}v_c & = (S_{uuv} + S_{vvv}) / 2,\\
\end{aligned}
\end{equation}
where $S_{uu} = \sum_j {(u_i^j-\bar{u})}^2$, $S_{vv} = \sum_j {(v_i^j-\bar{v})}^2$, $S_{uv} = \sum_j {(u_i^j-\bar{u})(v_i^j-\bar{v})}$, $S_{uuu} = \sum_j {(u_i^j-\bar{u})}^3$, $S_{vvv} = \sum_j {(v_i^j-\bar{v})}^3$, $S_{uuv} = \sum_j {(u_i^j-\bar{u})}^2(v_i^j-\bar{v})$, $S_{uvv} = \sum_j {(u_i^j-\bar{u})}(v_i^j-\bar{v})^2$, $\bar{u} = \frac{1}{n}\sum_j u_i^j$, $\bar{v} = \frac{1}{n}\sum_j v_i^j$. 
Finally, the circle $\mathcal{L}_i = (l_x, l_y, r)$ with position $(l_x, l_y)$ and radius $r$ is obtained by 
\begin{equation}
\begin{aligned}
\label{circle_center}
l_x & = u_c + \bar{u},\\
l_y & = v_c + \bar{v},\\
r & = \frac{1}{n}\sum_{j=1}^{n} \sqrt{(u_i^j-l_x)^2 + (v_i^j-l_y)^2},\\
\end{aligned}
\end{equation}
Here $r$ is directly calculated rather than the method in \cite{bullock2006least}.

To extract the semantic information, we obtain the feature vectors $\mathbf{F_i}$ = $\{\mathbf{f}_i^1,...,\mathbf{f}_i^n\}$, where $\mathbf{f}_i^j, j = 1, ..., n$ is a $d$-dimension embedding of each pixel  in the range-view image that belongs to the $i^{th}$ pole-like object. Then the feature vector $\mathcal{F}_i$ of this pole-like object is calculated by
\begin{equation}
\label{pole_feature}
\mathcal{F}_i = max(\mathbf{F_i}, 2),
\end{equation}
where $max(\mathbf{F_i}, 1)$ means the max operation for each column of $\mathbf{F_i}$ and $max(\mathbf{F_i}, 2)$ means the max operation for each row. The probability vector $\mathcal{C}_i$ and category $\hat{y}_i$ are calculated as
\begin{equation}
\label{pole_prob}
\mathcal{C}_i = \frac{1}{n}\sum_j \mathbf{c}_i^j,
\end{equation}
\begin{equation}
\label{pole_cls}
\hat{y}_i = \argmax \mathcal{C}_i,
\end{equation}
where $\argmax$ takes the index of the maximum value in probability vector $\mathcal{C}_i$.

\begin{figure*}[htbp]
    \centering
	\includegraphics[width=0.99\textwidth]{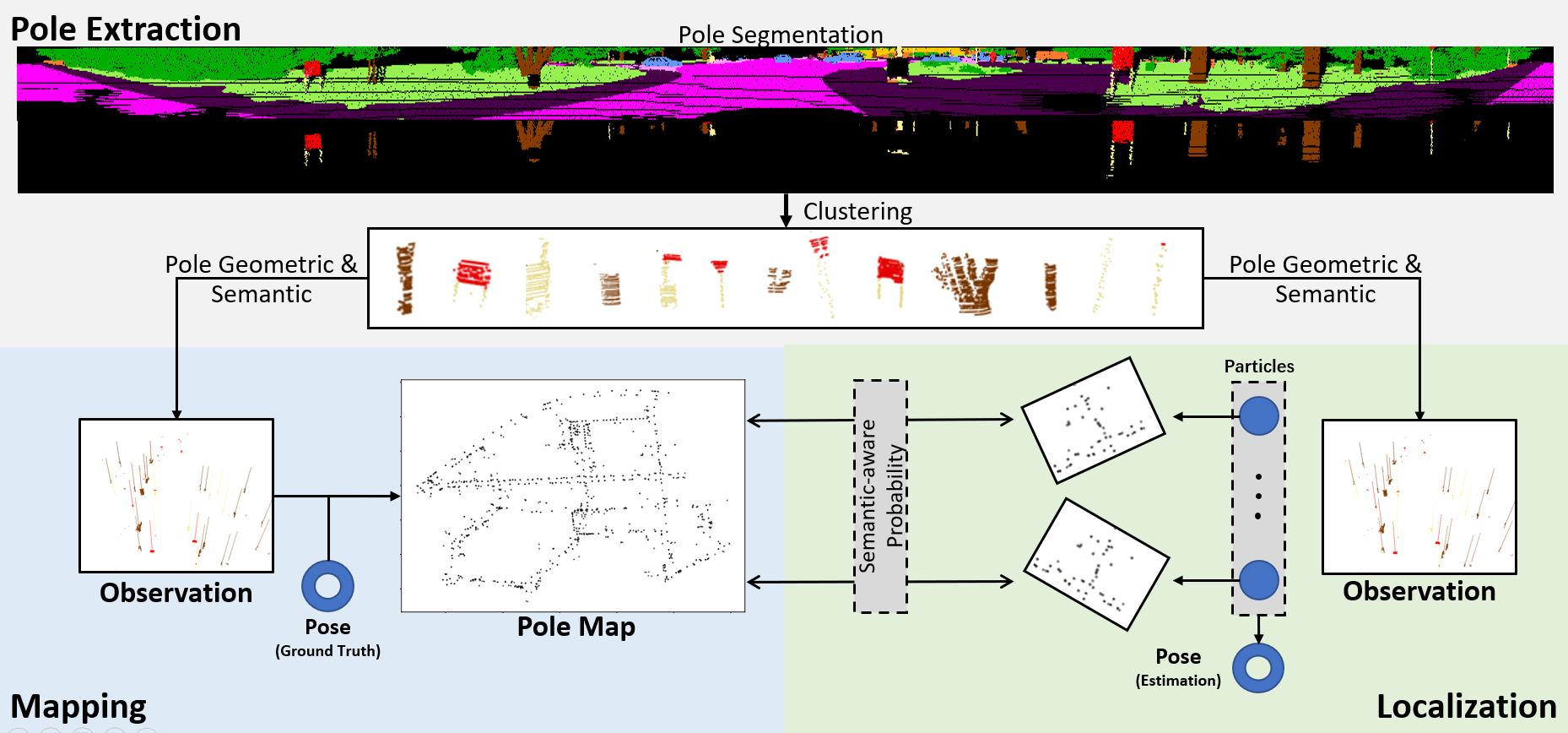} 
	\caption{The procedure of pole extraction, mapping, and localization. On the top is the pole extraction, the bottom left is the mapping, and the bottom right is the localization.}
	\label{architecture}
\end{figure*}

Finally, the geometric attribute $\mathcal{L}_i$ and semantic attributes $\mathcal{F}_i$, $\mathcal{C}_i$, $\hat{y}_i$ are used to describe the $i^{th}$ pole-like objects. The pole extraction procedure is shown in Fig. \ref{architecture}. The above-observed geometric and semantic information is used in both mapping and localization. In the next section, we will introduce the mapping for those pole objects.

\subsection{Creation of Multi-layer Semantic Pole-Map}
After extracting information for pole-like instances, they are represented as circles $\mathcal{L}_i$ with semantic information in each LiDAR scan. Then the pole-like instances from all scans in a sequence are integrated and added to the global map to build the semantic pole-map. 
As a pole-like object may be observed multiple times during mapping, the pole-like instances may duplicate if they are the observations of the same pole-like object. Thus, we utilize the ground-truth ego-motion to integrate the duplicates of each segmented instance. A final landmark on the map is the aggregation of these duplicates. Different from \cite{dong2021online,dong2023online}, we take the categories of pole-like instances into consideration for more robust mapping.

Specifically, we divide the instances into groups according to their categories $\hat{y}_i$. For each group, we aggregate the pole-like instances into clusters according to their connectivity. The two instances are deemed as connective if they overlap with each other. 
In this way, multiple clusters are generated for one pole-like object according to different categories from different observations.
We denote the $i^{th}$ cluster of circles, feature vectors, and probability vectors as $\mathcal{L}^c_i$ = $\{\mathcal{L}_1,...,\mathcal{L}_m\}$,
$\mathcal{F}^c_i$ = $\{\mathcal{F}_1,...,\mathcal{F}_m\}$,
$\mathcal{C}^c_i$ = $\{\mathcal{C}_1,...,\mathcal{C}_m\}$, 
respectively, where $m$ is the number of pole-like instances in this cluster.
The final circle $\mathcal{L}_i^g$, feature vector $\mathcal{F}_i^g$, probability vector $\mathcal{C}_i^g$ and category $\hat{y}_i^g$ of the $i^{th}$ pole-like landmark in map are the aggregation of instances in each cluster,
\begin{equation}
\begin{aligned}
\label{pole_desc}
\mathcal{L}_i^g &= \frac{1}{m}\sum_j \mathcal{L}_j, \mathcal{L}_j \in \mathcal{L}^c_i,\\
\mathcal{F}_i^g &= \frac{1}{m}\sum_j \mathcal{F}_j, \mathcal{F}_j \in \mathcal{F}^c_i,\\
\mathcal{C}_i^g &= \frac{1}{m}\sum_j \mathcal{C}_j, \mathcal{C}_j \in \mathcal{C}^c_i,\\
\hat{y}_i^g &= \argmax \mathcal{C}_i^g.\\
\end{aligned}
\end{equation}

\begin{figure}[t]
  \centering
  \includegraphics[width=0.99\linewidth]{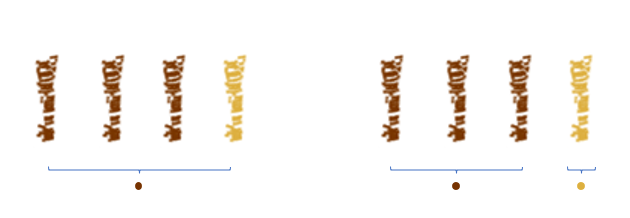}
  \caption{Left: The aggregation of observations of the same landmark without considering their semantics categories. When the observation is uncertain during pole segmentation and extraction, the aggregation of information from different categories into one landmark makes semantic features ambiguous (e.g., the trunk is categorized as a pole in one of the observations). 
  Right: The aggregation considering their semantics. Observations of different semantics are distinguished and aggregated into different layers (landmarks), which provides more robust geometric and semantic features.}
  \label{fig:pole_layer}
\end{figure}

The instances with different categories are aggregated into different clusters and generate different landmarks. In this way, a multi-layer pole-map is built, in which each layer contains pole-like landmarks for the same category. This is different from \cite{dong2021online,dong2023online} as we aggregate the instances in separated categories, as shown in Fig. \ref{fig:pole_layer}. 
When taking the average of geometric and semantic information of instances for each cluster, the multi-layer mapping reduces the mixture of ambiguous information in different categories from noisy pole segmentation and extraction. Thus, multi-layer mapping provides more robust landmarks in pole-maps. 
In the ablation study Section \nameref {sec:multilayermapping} we will investigate the superiority of multi-layer mapping. The mapping procedure is shown in Fig. \ref{architecture}.

\subsection{Semantic-Aware Particle Filtering}
In localization given the semantic pole-map, the observed pole-like objects are used to estimate the vehicle's ego-motion in the map (Fig. \ref{architecture}). We propose a semantic-aware particle-filter localization approach based on observing the pole-like structures to achieve robustness and accuracy of the localization. Conventionally, the general particle filter
has been widely used in localization in a map \citep{dellaert1999monte}.

The critical issues in particle-filter localization given a map are twofold. The first one is on data association, which establishes the correspondences between the pole-like objects from the online observation and those from the map. The association is unknown and should be created online. The second issue is on updating each particle's weight. 

The conventional methods in dealing with the first issue are based on  matching the nearest neighbors. 
However, the data association step usually contains erroneous correspondences due to the effect of dense dynamic participators, especially the presence of pole-like pedestrians during online localization or mapping. To deal with this issue, some methods have been proposed to incorporate semantic information into optimization for proper data association \citep{jiang2021particle,miller2021any,akai2020semantic,bavle2018stereo,zimmerman2022long}. 

For the second issue, the weight of each particle in the conventional particle filter-based method is estimated based on the probability of the observation conditioned on this particle's state \citep{dellaert1999monte}. Given the $k^{th}$ particle, assuming its weight at time $t-1$ is $w_{t-1}^k$, let $\textbf{s}_t^k$ be the state of the particle $k$ at time $t$. With observations $\textbf{O}_t = \{ o_t^1, ..., o_t^m \}$ at time $t$, the probability of $\textbf{O}_t$ conditioned on $\textbf{s}_t^k$ and map $\mathbf{M}$ can be calculated as
\begin{equation}
\begin{aligned}
\label{particle_measurement}
p_t^k & = \mathbb{P}(\textbf{O}_t \mid \textbf{s}_t^k;\mathbf{M}),\\
\end{aligned}
\end{equation}
Where $\mathbf{M}$ is the map consisting of landmark positions $l = (l_x, l_y)$. In the implementation, Eq. \ref{particle_measurement} can be realized as the product of Gaussian distributions assuming the independence of each observation,
\begin{equation}
\begin{aligned}
\label{particle_implementation}
p_t^k & = \mathbb{P}(D_t^k)\\
& = \prod\limits_{d_t^k(j) \in D_t^k} \mathbb{P}(d_t^k(j) \mid 0;\sigma^2)\\
& = \prod\limits_{d_t^k(j) \in D_t^k} \frac{1}{\sigma\sqrt{2\pi}}e^-{\frac{(d_t^k(j))^2}{2\sigma^2}}, j = 1, ..., m,
\end{aligned}
\end{equation}
where $D_t^k$ represents the positional discrepancy $\mathbb{D}$ between the actual observations $\textbf{O}_t$ and the expected ones made by the $k^{th}$ particle given the particle's state $\textbf{s}_t^k$ and map $\mathbf{M}$. Formally, $D_t^k$ is defined as
\begin{equation}
\begin{aligned}
\label{particle_distance}
D_t^k &= \mathbb{D}(\textbf{O}_t \mid \textbf{s}_t^k;\mathbf{M})\\
&= \{d_t^{k}(j) \mid d_t^{k}(j)=\min\limits_{l \in \mathbf{M}} \lVert l-\textbf{s}_t^k\textbf{o}_t^j \rVert_2, \textbf{o}_t^j \in \textbf{O}_t\},
\end{aligned}
\end{equation}
where $\textbf{s}_t^k\textbf{o}_t^j$ represents the estimated position of observation $\textbf{o}_t^j$ in the map given the $k^{th}$ particle's state $\textbf{s}_t^k$. The $\min\limits_{l \in \mathbf{M}} \lVert l-\textbf{s}_t^k\textbf{o}_t^j \rVert_2$ is used to measure the discrepancy between $\textbf{s}_t^k\textbf{o}_t^j$ and its neighbor through some data association strategies, where the nearest neighbor searching scheme is often used for simplicity.
Then the weight of the $k^{th}$ particle, $w_t^k$, is updated as
\begin{equation}
\label{particle_weight}
w_t^k = \frac{p_t^k}{\sum\limits_i p_t^i} w_{t-1}^{k},
\end{equation}

Although a few existing works have proposed to utilize semantic information to update particle weights \citep{jeong2020hdmi,bernuy2018topological,yan2019global}, 
different from the existing works, we utilize the semantic information of pole-like landmarks, extracted by the LiDAR sensor, which is efficient and robust to illumination.
In this work, the semantic attributes $\mathcal{F}_i$ and $\hat{y}_i$ in Eq. \ref{pole_feature} and Eq. \ref{pole_cls}, respectively, and $\mathcal{F}_i^g$ and $\hat{y}_i^g$ in Eq. \ref{pole_desc} are utilized to improve the landmark correspondence and particle weight calculation. With these semantic attributes, poles in different types can be distinguished, and the correspondence quality is taken into consideration. In the following, we introduce them in turn.

\begin{figure}[t]
  \centering
  \includegraphics[width=0.99\linewidth]{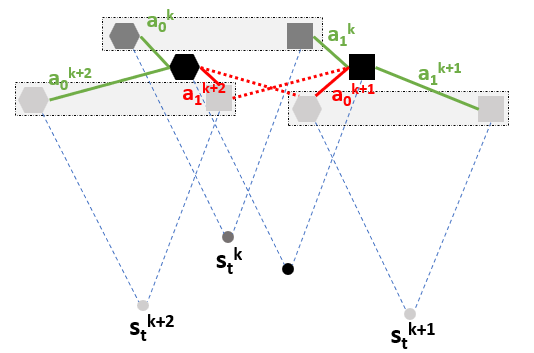}
  \caption{Illustration of why the nearest neighbor data association can result in an overestimation of the particle's updated weight, see the text for explanation. 
  }
  \label{fig:inconsistency}
\end{figure}

\subsubsection{Semantic-Aware Inconsistency}\label{semantic_ic}
\hfill

Following Eq. \ref{particle_distance}, the nearest-neighbor searching is used to find the correspondences for each particle. As shown in 
Fig. \ref{fig:inconsistency}, four particles (circular points) are sampled with one as the ground-truth vehicle pose (the dark circular point) and the other three ($\textbf{s}_t^k$, $\textbf{s}_t^{k+1}$, and $\textbf{s}_t^{k+2}$) are not at the vehicle's real location). 

The expected landmark observations by the three particles are shown in the grey rectangles, respectively. The dark hexagon and square are two landmarks on the map. The associations of the expected landmarks of $\textbf{s}_t^k$, $\textbf{s}_t^{k+1}$, and $\textbf{s}_t^{k+2}$ to the map landmarks are obtained by the nearest neighbor searching and shown as the red and green solid lines. Obviously, the red solid lines correspond to the wrong associations and the true associations should be the ones represented by the green solid lines and red dotted ones. 
In this example, we can observe that the distance between the wrong associated landmarks is smaller than that of the true associations. 
Therefore, based on Eq. \ref{particle_distance}, the nearest-neighbor searching can result in wrong particle weights estimation, i.e., an over-estimation of a particle's weight which is supposed to have a low weight.

For this problem, formally let us suppose that we have $m$ landmark observations in the map at time $t$ with the ground truth correspondences
\begin{equation}
\begin{aligned}
\label{gt_corr}
\bar{A} = \{\bar{a}_0, \bar{a}_1, ..., \bar{a}_m\}
\end{aligned}
\end{equation}
and the approximated correspondences by the $k^{th}$ particle (based on nearest-neighbor search Eq. \ref{particle_distance}) 
\begin{equation}
\begin{aligned}
\label{pd_corr}
{A}^k = \{{a}_0^k, {a}_1^k, ..., {a}_m^k\}
\end{aligned}
\end{equation}
where $\bar{a}_i$ and ${a}_i^k$ are the indices of pole-like landmarks in map $\mathbf{M}$, we have
\begin{equation}
\begin{aligned}
\label{particle_association}
\bar{D}_t^k = \{\bar{d}_t^k(j) \mid \bar{d}_t^k(j) = \lVert \mathbf{M}_{\bar{a}_j}-\textbf{s}_t^k\textbf{o}_t^j \rVert_2, j = 1, ..., m\},\\
{D}_t^k = \{d_t^{k}(j) \mid d_t^{k}(j) = \lVert \mathbf{M}_{{a}_j^k}-\textbf{s}_t^k\textbf{o}_t^j \rVert_2, j = 1, ..., m\},\\
\end{aligned}
\end{equation}
where $\mathbf{M}_\#$ is the location of the $\#^{th}$ landmark in the map $\mathbf{M}$.
The ground truth correspondence $\bar{A}$ is independent of particle state, while the approximated correspondence ${A}^k$ is dependent on a specific particle state when using the nearest neighbor scheme to find the approximate correspondence such as in Eq. \ref{particle_distance}.

From the above analysis with Fig. \ref{fig:inconsistency}, we know
\begin{equation}
\begin{aligned}
\label{particle_distance_inequation_bar}
\lVert \mathbf{M}_{{a}_j^k}-\textbf{s}_t^k\textbf{o}_t^j \rVert_2 & \leq \lVert \mathbf{M}_{\bar{a}_j}-\textbf{s}_t^k\textbf{o}_t^j \rVert_2,\\
{d}_t^k(j) & \le \bar{d}_t^k(j),\\
\end{aligned}
\end{equation}
Based on Eq. \ref{particle_implementation}, thus the probability $\bar{p}_t^k$ given ground truth correspondence $\bar{A}$ and the probability $p_t^k$ given approximated correspondence $A^k$ hold
\begin{equation}
\begin{aligned}
\label{particle_prob_inequation_bar}
{p}_t^k \ge \bar{p}_t^k.\\
\end{aligned}
\end{equation}
The equality holds when the approximated correspondence of the particle is the same as the ground truth correspondence, which can be achieved when the particle's state is the same as or close to the vehicle's ground-truth state (shown by the particle $\textbf{s}_t^k$ from Fig. \ref{fig:inconsistency}).
However, for the particles whose states are not close to ground-truth poses (shown by the particle $\textbf{s}_t^{k+1}$ and $\textbf{s}_t^{k+2}$ in Fig. \ref{fig:inconsistency}), the inequality holds, which brings higher weights for these particles than expected.
 Thus, the equality holds for those "good" particles $\textbf{s}_t^k$ and the inequality holds for those "bad" particles $\textbf{s}_t^{k+1}$ and $\textbf{s}_t^{k+2}$, which means large weights are assigned to those "bad" particle and hinders the convergence and incurs degraded localization performance.

To address this problem, we incorporate the data-association quality of $k^{th}$ particle into the calculation of particle weights. Given the semantic observation $\mathbf{O^s}_{t}$ consisting of $\mathcal{F}_i$ and $\hat{y}_i$, and map $\mathbf{M^s}$ consisting of $\mathcal{F}_i^g$ and $\hat{y}_i^g$, we define semantic-aware correspondence inconsistency $I_t^k$ to evaluate the data-association quality as
\begin{equation}
\label{particle_inconsitency}
I_t^k = \mathbb{I}(A^k \mid \mathbf{O^s}_{t},\mathbf{M^s}),
\end{equation}
where $\mathbb{I}$ will be introduced later. Then the probability of particle $\textbf{s}_t^k$ in Eq. \ref{particle_measurement} is replaced by
\begin{equation}
\begin{aligned}
\label{particle_measurement_union}
\tilde{p}_t^k
& = \mathbb{P}(\mathbf{O}_{t}, \mathbf{O^s}_{t}  \mid \textbf{s}_t^k;\mathbf{M};\mathbf{M^s}) \\
& = \mathbb{P}(\mathbf{O}_{t} \mid \textbf{s}_t^k;\mathbf{M}) \mathbb{P}(\mathbf{O^s}_{t} \mid \textbf{s}_t^k;\mathbf{M^s}), \\
\end{aligned}
\end{equation}
where $\mathbf{O}_{t}$ and $\mathbf{O^s}_{t}$ are assumed to be independent. Because the condition probability of $\mathbf{O^s}_{t}$ is only dependent on $A^k$ (decided by $\textbf{s}_t^k$) and $\mathbf{M^s}$, $\tilde{p}_t^k$ can be reduced to
\begin{equation}
\begin{aligned}
\label{particle_measurement_union2}
\tilde{p}_t^k
& = p_t^k \mathbb{P}(\mathbf{O^s}_{t} \mid A^k;\mathbf{M^s}). \\
\end{aligned}
\end{equation}
Different from $\textbf{s}_t^k$ which is unique for all particles, the ${A}^k$ from different particles can be the same as long as their correspondences are the same. Supposing the function $\mathbb{I}$ returns $0$ for correct correspondences and $1$ for wrong correspondences. 
As in Eq. \ref{particle_implementation}, Eq. \ref{particle_measurement_union2} can be realized as the product of Gaussian distributions assuming the independence of each observation,
\begin{equation}
\begin{aligned}
\label{particle_measurement_counter}
\tilde{p}_t^k
& = p_t^k \mathbb{P}(I_t^k)\\
& = p_t^k \prod\limits_{i}^{N_t} \mathbb{P}(0 \mid 0;\sigma^2) \prod\limits_{i}^{N_f} \mathbb{P}(1 \mid 0;\sigma^2) \\
& \le p_t^k \prod\limits_{i}^{N_t+N_f} \mathbb{P}(0 \mid 0;\sigma^2) \\
& = p_t^k \mathbb{P}(\mathbf{O^s}_{t} \mid \bar{A};\mathbf{M^s}). \\
\end{aligned}
\end{equation}
where $I_t^k$ represents the correspondence inconsistency between
the ground-truth semantic observations $\mathbf{O^s}_{t}$ and the expected ones made by
the $k^{th}$ particle given the correspondence ${A}^k$ and the semantic pole-map $\mathbf{M^s}$. $N_t$ and $N_f$ are the numbers of correct and wrong correspondences, respectively. Again, because the approximation of correspondence is accurate for the particles close to ground truth, the equality holds for those ”good” particles and the inequality holds for those ”bad” particles, which means smaller weights are assigned to those "bad" particles and improves the localization performance.


In Eq. \ref{particle_measurement_counter}, we need the correspondence inconsistency evaluation function $\mathbb{I}$ to update $\tilde{p}_t^k$ . Therefore, we propose to use the semantic discrepancy to realize function $\mathbb{I}$ as
\begin{equation}
\label{particle_semantic_inconsitency_1}
\mathbb{I}(A^k) = \{1 - cos(\mathcal{F}^g_{{a}_i^k}, \mathcal{F}_{i}) \mid i \in (0, 1, ..., n)\},
\end{equation}
where $cos()$ is the cosine similarity and ${a}_i^k$ is the index of associated landmark in $\mathbf{M^s}$ of the $i^{th}$ observation given the $k^{th}$ particle. In Eq. \ref{particle_semantic_inconsitency_1}, when the $d$-dimension feature vector $\mathcal{F}^g_{{a}_i^k}$ and $\mathcal{F}_{i}$ are close, their inconsistency is small and they have high possibility to be the same pole-like objects, which can be used to approximate the function $\mathbb{I}$. 

\subsubsection{Semantic-Aware Nearest Neighbor}\label{semantic_nn}
\hfill

\begin{figure}[t]
  \centering
  \includegraphics[width=0.99\linewidth]{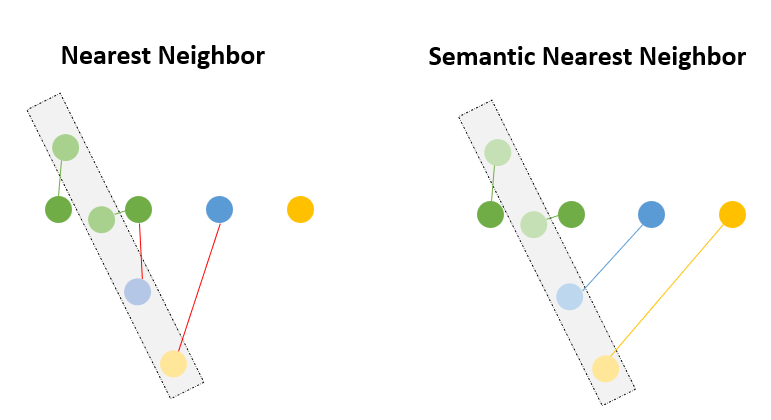}
  \caption{Left: the nearest neighbor searching results in wrong correspondences. Right: the semantic nearest neighbor searching finds the correct correspondences. The light-colored nodes in the rectangle are the observations, and the dark-colored nodes out of the rectangle are the landmarks in the map. Different colors of nodes represent different semantic categories, and the red edges represent the wrong correspondences.}
  \label{fig:semcor}
\end{figure}

To further improve the correspondence quality, the semantic categories of these pole-like landmarks are used, as shown in Fig. \ref{fig:semcor}. We replace $\mathbf{M}$ in Eq. \ref{particle_distance} by its subset $\mathbf{M}' = \{l_j \mid l_j \in \mathbf{M}, \hat{y}_j = \hat{y}_i\}$, where $\hat{y}_i$ is the predicted semantic category of the $i^{th}$ observed pole-like landmark and $\hat{y}^g_j$ of the $j^{th}$ pole-like landmark in map, we have
\begin{equation}
\begin{aligned}
\label{particle_distance_prime}
D_t^{\prime k} &= \mathbb{D}^\prime(\textbf{O}_t \mid \textbf{s}_t^k;\mathbf{M^\prime})\\
&= \{d_t^{\prime k}(j) \mid d_t^{\prime k}(j)=\min\limits_{l^\prime \in \mathbf{M}^\prime} \lVert l^\prime-\textbf{o}_t^j\textbf{s}_t^k \rVert_2, \textbf{o}_t^j \in \textbf{O}_t^j\},
\end{aligned}
\end{equation}
This way, the correspondences between the pole-like landmarks in observation and in the map are restricted to the same semantic category. 
Comparing $D_t^{\prime k}$ with $D_t^k$ will have,
\begin{equation}
\begin{aligned}
\label{particle_distance_improve_prime}
\min\limits_{l' \in \mathbf{M}'} \lVert l'-\textbf{o}_t^j\textbf{s}_t^k \rVert_2 & \ge \min\limits_{l \in \mathbf{M}} \lVert l-\textbf{o}_t^j\textbf{s}_t^k \rVert_2,\\
{d}_t^{\prime k}(j) & \ge d_t^k(j),\\
\end{aligned}
\end{equation}
and thus ${p}_t^{\prime k}$ holds
\begin{equation}
\begin{aligned}
\label{particle_prob_improve_prime}
{p}_t^{\prime k} \leq {p}_t^k.\\
\end{aligned}
\end{equation}
Similar to Eq. \ref{particle_distance_inequation_bar} and \ref{particle_prob_inequation_bar}, in one of the $K$ classes for pole-like objects, the nearest neighbor is always closer than other neighbors, including the neighbor representing the ground truth correspondence, we have,
\begin{equation}
\begin{aligned}
\label{particle_distance_inequation_prime}
{d}_t^{\prime k}(j) & \le \bar{d}_t^k(j),\\
\end{aligned}
\end{equation}
\begin{equation}
\begin{aligned}
\label{particle_prob_inequation_prime}
{p}_t^{\prime k} \ge \bar{p}_t^k,\\
\end{aligned}
\end{equation}
and combined with Eq. \ref{particle_distance_improve_prime} and \ref{particle_prob_improve_prime} we have,
\begin{equation}
\begin{aligned}
\label{particle_distance_improve_inequation_prime}
d_t^k(j) & \le {d}_t^{\prime k}(j) \le \bar{d}_t^k(j),\\
\end{aligned}
\end{equation}
\begin{equation}
\begin{aligned}
\label{particle_prob_improve_inequation_prime}
{p}_t^k & \ge {p}_t^{\prime k} \ge \bar{p}_t^k.\\
\end{aligned}
\end{equation}
This means that the probability of a particle with semantic data association (${p}_t^{\prime k}$) is always smaller than or equal to that with the normal data association (${p}_t^k$), and is always closer to the expected probability derived from ground truth correspondences than the normal data association.
The equality holds for the particles whose poses are close to the true poses when both the nearest neighbor and the semantic nearest neighbor searching find the true correspondences. The inequality holds for the particles whose poses are far from the true ones when the semantic nearest neighbor searching finds better correspondences than the nearest neighbor searching. 
In another word, the weights of these "bad" particles are reduced by finding a further correspondence. Simply speaking, the proposed semantic-aware nearest neighbor approach to the true correspondences using semantic information, which improves the localization performance.

\section{Experiments}  
\label{sec:exp}

\subsection{Datasets}
We use SemanticKITTI \citep{behley2019semantickitti} to validate our methods. SemanticKITTI provides 43551 LiDAR scans collected by Velodyne-HDLE64 LiDAR in 22 sequences. 
To evaluate semantic segmentation performance, sequences 00-10 except 08 are used as the training set and sequence 08 as the validation set.
The semantic labels are available in the training and validation set. Besides, it provides ground truth poses estimated by \cite{behley2018efficient} for all sequences. Localization performance is evaluated in sequences 11-21 to show its generalization ability.

\subsection{Pole Segmentation}
\textbf{Metrics.} We use mIoU \citep{everingham2015pascal} as the metric for the semantic segmentation of pole-like objects, by which the higher mIoU the more overlap between prediction and ground truth. 
We use sequences 00-10 except 08 to train the segmentation model and to evaluate the model on sequence 08. 
We use SalsaNext \citep{cortinhal2020salsanext} as the backbone in the mask-classification paradigm. We use all the categories, e.g., road, car, and building, for training, and focus the performance on $3$ pole-like categories, i.e., pole, trunk, and traffic sign. These $3$ categories are used to build the semantic pole-map.
 
During training and inference, the point cloud is projected into a range view with a resolution of $64*2048$. Although there is information loss after projection, i.e., some points are out of view and some points are occluded by the ones in front of them, we do not re-project the segmentation result back to the original point cloud, so there is no leaking problem as in \cite{milioto2019rangenet++}. This means we only predict the semantics for those points available in the range-view representation instead of all points and these points are enough for feature learning and pole-map construction. We compare ours with SalsaNext \citep{cortinhal2020salsanext} and FIDNet \citep{zhao2021fidnet} in evaluating semantic segmentation performance, as they are semantic segmentation methods based on the range-view-based representation for LiDAR scans. As shown in Table \ref{pole_miou}, we can see that our model trained by the mask-classification paradigm with data augmentation achieves the best performance. 

\begin{table}[t]
\caption[Pole Segmentation]{The semantic segmentation performance of three types of pole-like landmarks on SemanticKITTI \citep{behley2019semantickitti} validation set. $\dagger$ represents the evaluation of points that are available in the range-view representation instead of re-projecting them back to the original point cloud. Experiments without $\dagger$ represent the evaluation of all points by re-projecting points available in the range-view representation back to the original point cloud.}
\Huge
\resizebox{0.98\columnwidth}{!}{
\begin{tabular}
{|l|c|cccc|}
\toprule
Method
& \text{Mean IoU} 
& \text{Pole} 
& \text{Trunk}
& \text{Traffic-sign}
& \text{Others}\\
\midrule
\midrule
SalsaNext \citep{cortinhal2020salsanext}$\dagger$
& 59.7 & 58.7 & 63.9 & 44.9 & 60.4 \\
\midrule
FIDNet \citep{zhao2021fidnet}$\dagger$
& 60.4 & 60.1 & 68.0 & 44.1 & 61.0 \\
\midrule
SalsaNext w/ WPD$\dagger$
& 63.4 & 64.3 & 66.7 & 49.3 & 64.0 \\
\midrule
Ours w/o WPD$\dagger$
& 59.1 & 64.5 & 66.9 & 45.8 & 59.1 \\
\midrule
Ours$\dagger$
& 66.4 & 65.0 & 67.5 & 48.4 & 67.5 \\
\midrule
\midrule
SalsaNext \citep{cortinhal2020salsanext}
& 57.5 & 56.7 & 60.8 & 44.9 & 58.1 \\
\midrule
FIDNet \citep{zhao2021fidnet}
& 58.8 & 57.6 & 64.0 & 43.7 & 59.5 \\
\midrule
SalsaNext w/ WPD
& 60.9 & 61.3 & 62.4 & 48.1 & 61.6 \\
\midrule
Ours w/o WPD
& 57.2 & 61.5 & 62.7 & 45.3 & 57.3 \\
\midrule
Ours
& 63.7 & 61.9 & 63.2 & 47.6 & 64.9 \\
\bottomrule
\end{tabular}
}
\label{pole_miou}
\end{table}

\begin{figure*}[htbp]
    \centering
	\includegraphics[width=0.99\textwidth]{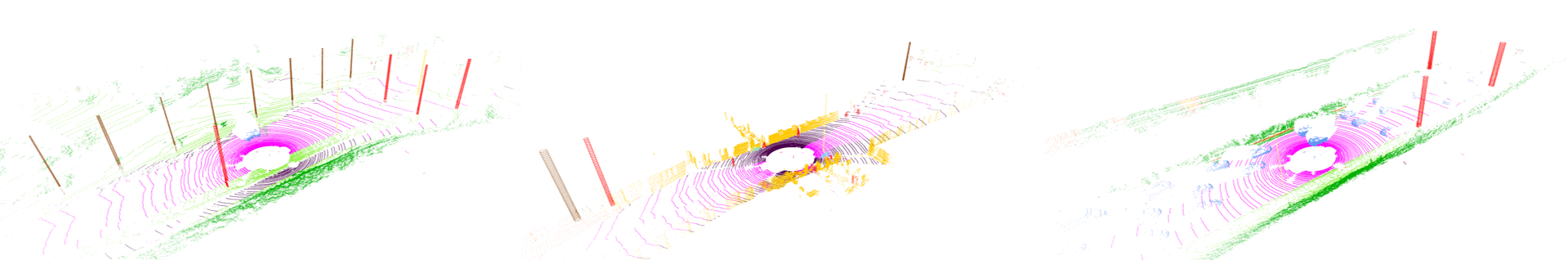} 
	\caption{The visualization of LiDAR semantic segmentation of poles. The color of each LiDAR point represents the predicted semantic label. The cylinders represent the poles extracted from the predicted semantic labels with categories denoted by color and radius denoted by thickness. Note that we also show the semantics of some other objects such as roads for better visualization.}
	\label{fig:polevis}   
\end{figure*}

Fig. \ref{fig:polevis} shows three examples of LiDAR semantic segmentation with extracted poles. The color of each LiDAR point represents the predicted semantic label. The cylinders represent the poles extracted from the predicted semantic labels with categories denoted by color and radius denoted by thickness. 
On the left, middle, and right show the scenes with regular plated trees, pole-like pedestrians, dense dynamic participators on the road or roadside. As shown on the left, our method accurately extracts the pole-like structures, i.e., trees on the roadside. As shown on the middle, no pole-like pedestrians have been misclassified as poles. As shown on the right, pole-like structures, i.e., traffic-signs on the highway, can be localized among dense dynamic participators.
Next, we will analyze the mapping performance based on semantic segmentation results.

\subsection{Pole-Map Creation}\label{exp:mapping}
\textbf{Metrics.} We use the F1 score \citep{sokolova2009systematic} to evaluate the precision of our pole-map. The F1 score is the combination of recall and precision for pole-like objects. A pole-like object is regarded as a True Positive (TP) if the distance between the detected pole and its nearest neighbor in the ground truth map is no larger than $1$ meter. Otherwise, it is a False Positive (FP). The poles that are present in the ground truth map but not predicted are regarded as False Negatives (FN). The recall $R$, precision $P$, and F1 score $F$ is calculated as:
\begin{equation}
\begin{aligned}
\label{F1}
R & = N_{TP} / (N_{TP} + N_{FN}),\\
P & = N_{TP} / (N_{TP} + N_{FP}),\\
F & = 2*R*P/(R+P),
\end{aligned}
\end{equation}
where $N_{TP}$, $N_{FP}$, $N_{FN}$ is th number of TPs, FPs, FNs.

To build our pole-map, we use the ground truth ego-poses provided by the SemanticKITTI dataset. Specifically, we use the frames every $\delta_d$ meters (selected as the keyframes) to build our pole-map for each sequence, i.e., the pole-like objects are segmented in each keyframe and aggregated based on their locations and the ground-truth ego-poses of the vehicle. 
$\delta_d$ is set to $10$ in our experiments, as the position difference of two consecutive frames between mapping and localization is $5$ meters on average, which is large enough to evaluate the localization performance.
The different choices of $\delta_d$ will be investigated in Section \nameref{sec:mappingdist}. 

Because the depth of objects beyond 50 meters is generally quite noisy, pole-like objects beyond the range of 50 meters are ignored. 
To evaluate our pole-map accuracy, we compare it with the ground truth pole-map provided by \cite{dong2023online}, which is built from the ground-truth semantic labels in SemanticKITTI. 
In SemanticKITTI sequences 00-10, sequence 08 is usually used as the validation set and the remaining sequences are used as the training set. In addition to this splitting, we also use sequence 01 as the validation set and the remaining as the training set to further demonstrate the generalization ability. 

The comparison is shown in Table \ref{pole_f1}. We compare our pole-map accuracy with that of \cite{dong2021online,dong2023online} which created their own pole-map based on their pole detection results. Although the maps in \cite{dong2021online,dong2023online} are built with many more frames than ours in each sequence, our method outperforms them by a large margin. Because the codes of the learning-based method in \cite{dong2023online} is not yet released, we only fine-tune the hyperparameters in \cite{dong2021online} to build the map with our defined dataset as the baseline method \citep{dong2021online}$\dagger\dagger$. Besides, the accuracy of pole-map built with our baseline segmentation method \citep{cortinhal2020salsanext} is also compared. Among those methods, our model achieves the best performance in mapping.

\begin{figure*}[t]
  \centering
  \includegraphics[width=0.99\linewidth]{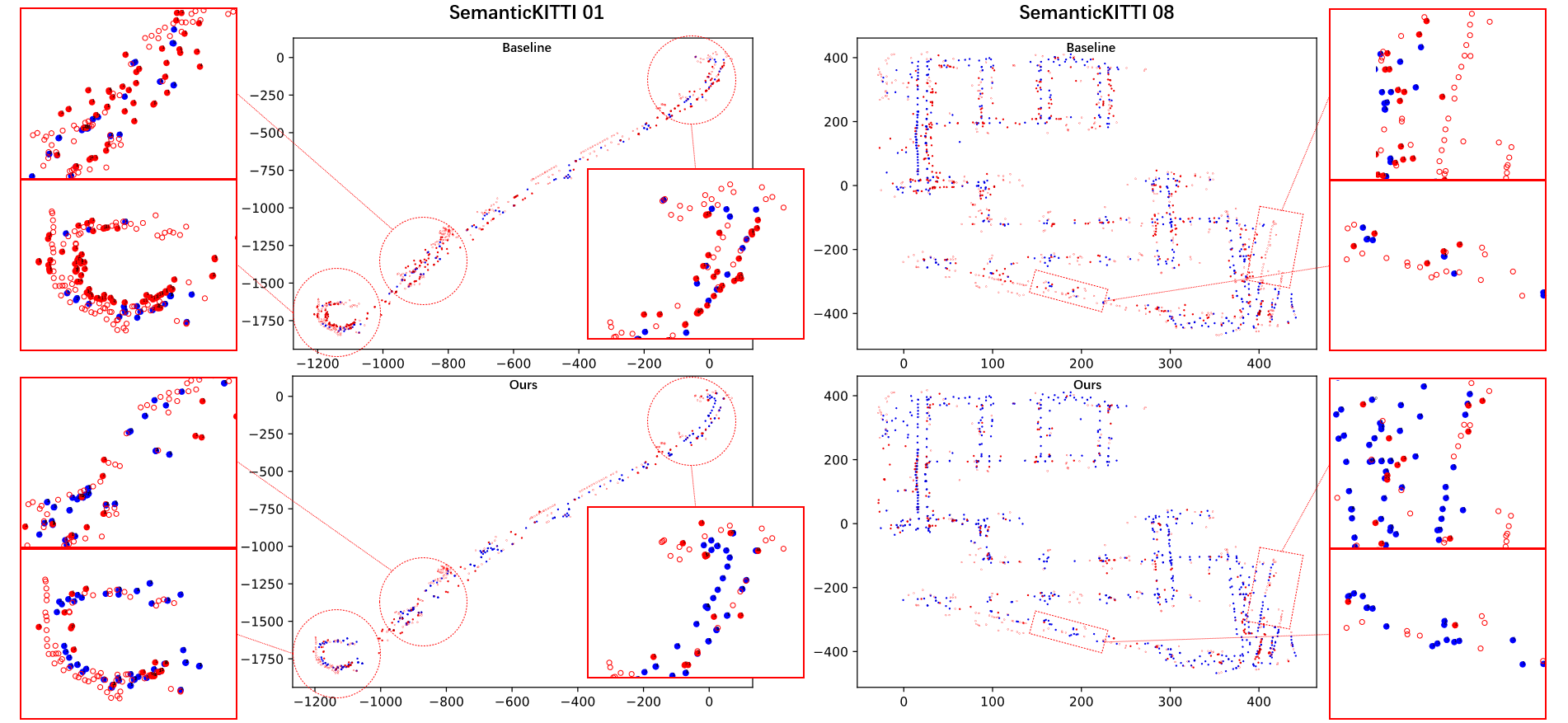}
  \caption{Visualization of pole-like maps for SemanticKITTI sequences 01 and 08. The top row shows the maps built by the baseline method \citep{dong2021online} tuned based on their published code, and the bottom shows our pole-maps. The blue and red points represent the pole-like objects in the built map, where the blue points are the true positives (TP) and the red points are the false positives (FP) by comparing with the ground-truth map provided by \cite{dong2023online}. The light-red circles represent the false negatives (FN). Our method has fewer FPs and FNs compared with the baseline method.}
  \label{fig:mapping}
\end{figure*}

Table \ref{pole_number} shows that our method can detect more poles on average, $N_s$, from a LiDAR scan than the baseline method \citep{dong2021online}$\dagger\dagger$,  contributing to the increase of an average number of observations $N_o$ of a pole, and not largely increasing the number of poles $N_m$ in a map. It shows that our segmentation method obtains more consistent pole extraction results.
Fig. \ref{fig:mapping} shows that the baseline method tends to predict more pole-like objects in sequence 01 and less in sequence 08, producing more false positives and false negatives. In contrast, our method is much higher in precision and recall. Next, we will compare the localization performance to show the importance of accurate pole-map creation.

\begin{table}[t]
\caption[Pole-map creation]{The comparison of mapping performance on the SemanticKITTI \citep{behley2019semantickitti} sequences 01 and 08. $\dagger$ the mapping results  reported at \cite{dong2021online,dong2023online} where the maps are built with more dense frames than ours in each sequence. $\dagger\dagger$  the baseline method tuned based on the published code in \cite{dong2021online}.}
\resizebox{0.98\columnwidth}{!}{
\begin{tabular}
{|l|c|c|c|c|c|c|}
\toprule
Sequence (validation)
& \multicolumn{3}{c|}{01}
& \multicolumn{3}{c|}{08}\\
\midrule
Method
& Precision
& Recall
& F1
& Precision
& Recall
& F1\\
\midrule
\citep{dong2021online}$\dagger$
& 0.52 & 0.12 & 0.19 & 0.78 & 0.41 & 0.54 \\
\citep{dong2023online}$\dagger$
& 0.67 & 0.27 & 0.38 & 0.60 & 0.59 & 0.60 \\
\midrule
Baseline \citep{dong2021online}$\dagger\dagger$
& 0.36 & 0.29 & 0.32 & 0.58 & 0.61 & 0.59 \\
SalsaNext \citep{cortinhal2020salsanext} 
& 0.49 & 0.78 & 0.60 & 0.64 & 0.94 & 0.76 \\
Ours
& 0.73 & 0.56 & 0.63 & 0.76 & 0.86 & 0.81 \\
\bottomrule
\end{tabular}
}
\label{pole_f1}
\end{table}

\begin{table}[t]
\caption[Pole-map creation]{The comparison of the number of poles $N_m$ in a map, the average number of poles $N_s$ for a LiDAR scan, and the average number of observations $N_o$ of a pole in SemanticKITTI \citep{behley2019semantickitti} sequence 01 and 08. $\dagger\dagger$ the baseline method tuned
based on the published code in \cite{dong2021online}.}
\resizebox{0.98\columnwidth}{!}{
\begin{tabular}
{|l|c|c|c|c|c|c|}
\toprule
Sequence (validation)
& \multicolumn{3}{c|}{01}
& \multicolumn{3}{c|}{08}\\
\midrule
Method
& $N_m$
& $N_s$
& $N_o$
& $N_m$
& $N_s$
& $N_o$\\
\midrule
Baseline \citep{dong2021online}$\dagger\dagger$
& 339 & 1.99 & 1.45 & 895 & 4.84 & 1.73 \\
Ours
& 325 & 3.21 & 2.44 & 960 & 10.08 & 3.37 \\
\bottomrule
\end{tabular}
}
\label{pole_number}
\end{table}

\subsection{Pole Localization}
\label{sec:localization}
\textbf{Metrics.} We calculate the mean absolute errors ($\Delta$) and root-mean-squared errors (RMSE) on position and heading to evaluate the performance of localization. These metrics represent the distance in meters from the prediction to the ground truth position and the difference in angle between the prediction and ground truth heading. The smaller these metrics the better the performance.

\textbf{Dataset.} To investigate the effectiveness of semantics information in localization, we evaluate the performance in the SemanticKITTI dataset.
Different from pole segmentation and pole-map creation which are evaluated in SemanticKITTI validation sequences 01 and 08, localization performance is evaluated in sequences 11-21 of the SemanticKITTI dataset to show its generalization ability. 

In the SemanticKITTI dataset,
the vehicle mostly drives in each specified route only once. Therefore, we do not have enough repeated routes to test the localization performance. 
Thus, we generate data suitable for location purposes from the original SemanticKITTI dataset to evaluate our method's localization performance.  
Our expectation on the data for localization is that the generated data should be as close as possible to the real ones that are collected when a vehicle  drives in the same scene at least twice, and the data used for localization should be different from the ones used for mapping. 

\begin{figure}[t]
  \centering
  \includegraphics[width=0.99\linewidth]{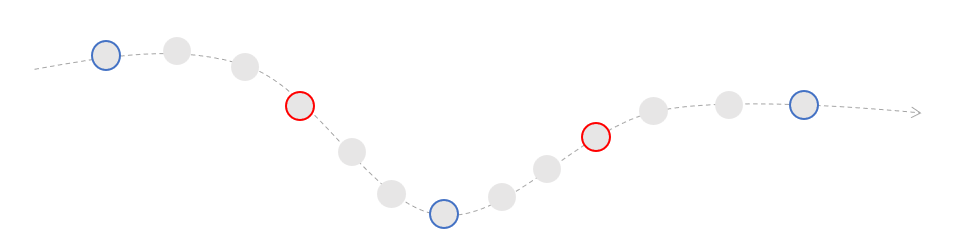}
  \caption{The illustration of data separation used for pole-map creation and localization test, respectively, from the same original SemanticKITTI sequence. Two subsets of frames are generated for each sequence. The pole-map corresponding to each sequence is built based on the selected keyframes (blue) along the route every $\delta_d$ meters according to the ground-truth ego-poses. Correspondingly, the localization test set is created by selecting those frames  (red) which are just in the middle way of two consecutive keyframes used for creating the pole-map. }
  \label{fig:keyframe}
\end{figure}

For that, we build our pole-maps with the  keyframes based on the ground-truth ego-poses with the keyframes sampled every $\delta_d$ meters along the route.
Our localization data is generated by selecting those frames which are just in the middle of two consecutive keyframes used for creating the pole-map.  
Fig. \ref{fig:keyframe} illustrates how we separate each original sequence of the SemanticKITTI dataset for pole-map creation and localization purposes, respectively.
In this way, two subsets of frames of the same original SemanticKITTI sequence are separated without overlap for mapping and localization, respectively.

\begin{figure*}[t]
  \centering
  \includegraphics[width=0.99\linewidth]{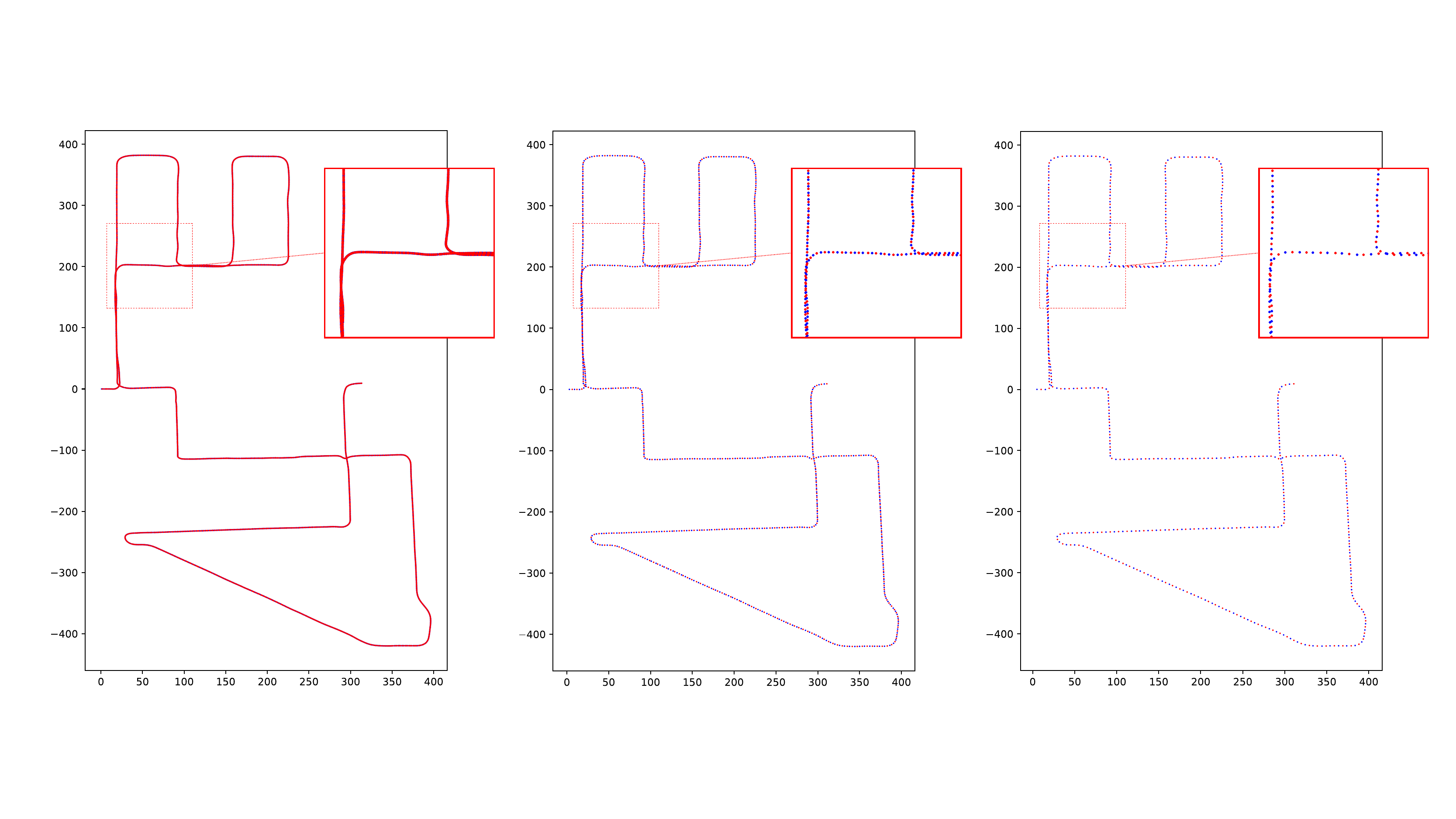}
  \caption{The examples of subsets of frames from SemanticKITTI sequence 08 using different $\delta_d$. Left, middle and right correspond to $1.5$, $6$, and $10$ meters for $\delta_d$, respectively. The blue points represent the frames used in mapping, and the red points represent the frames used in localization.}
  \label{fig:keyframe_example}
\end{figure*}

Fig. \ref{fig:keyframe_example} shows the examples of subsets of frames from SemanticKITTI sequence 08 using different $\delta_d$, in which the sequence is separated when $\delta_d$ is set to $6$ and $10$, respectively. Generally, the localization based on pole-like landmarks is difficult in scenarios where the pole-like landmarks are absent or very sparse because there are no enough landmarks to support the update of each particle's weight. 
Besides, the larger the accumulated error from the motion model that is used for predicting a particle's next movement, the harder the particle convergences.
From this point of view, 
the localization task is more difficult with a larger $\delta_d$ because the frames available for mapping are fewer and more accumulated errors will occur between two localization frames with a longer span, and the frames used for localization are more different from frames for mapping. If the motion model used for particle pose prediction has  
high uncertainty, localization will be even harder. 
The choice of $\delta_d$ will be investigated in Section \nameref{sec:mappingdist}. 

\textbf{Uncertainty.} 
In a practical particle-filter based localization framework given the map and odometry prediction (from wheel-odometer or lidar/camera-based odometry), the localization errors are mainly from 
the uncertainty of odometry prediction and noise of online observation. 

To test the robustness of our localization performance against the two types of uncertainty interruption, we simulate these two kinds of noise as follows. 
To simulate the odometry uncertainty, we add random noise of different degrees to the ground-truth poses provided by SemanticKITTI to generate the noisy odometry. 
Specifically, the odometry between two timestamps during localization is the odometry sampled from the multivariate normal distribution governed by the ground truth odometry as the mean and the covariance decided by $\phi_{odo}$. For the simulation of noisy online observations, we random drop $\phi_{O}$ percent of the poles detected online during localization to mimic the scenario of missing landmarks. 


\textbf{Compared methods.} We compare our localization performance with different pole-maps and different localization methods. 
\cite{dong2021online} proposed to extract the poles in the LiDAR's range-view representation by geometric rules, used the ground-truth ego-poses for mapping, and then did particle filter localization in the built map. \cite{wang2021pole} proposed to build the pole-map with the segmentation of pole-like landmarks based on RangeNet++ \citep{milioto2019rangenet++} and did the localization with iterative closest point (ICP) \citep{besl1992method} based on these pole-like landmarks.
We fine-tuned their methods to build the pole-map with our defined dataset and localize in the pole-map with simulation noise as described above. Then we compare the localization performance of our approaches with theirs. Finally, we use our pole-map to investigate the effectiveness of incorporating semantic information into the localization based on particle filtering.

\begin{table*}[t]
\caption[Pole Localization]{The comparison of localization performance in our semantic pole-maps and the 
maps of \cite{dong2021online,wang2021pole} created based on 
SemanticKITTI sequences 11-21. In our pole-map,  the normal particle filter localization scheme \citep{dellaert1999monte} and our semantic particle filter localization are compared. \textbf{PF} denotes the normal particle filter localization \citep{dellaert1999monte}. \textbf{I-PF} denotes semantic particle filter localization by incorporating the Semantic-Aware Inconsistency. 
\textbf{I+N-PF} denotes semantic particle filter localization by incorporating the Semantic-Aware Inconsistency and the Semantic-Aware Nearest Neighbor. In this experiment, $\phi_{odo}$ is set to $40$\%. $\phi_{O}$ is set to $80$\% in our methods and \cite{dong2021online}, and $0$\% in \cite{wang2021pole}. For results of  \cite{wang2021pole}, $*$ means that the localization fails across the whole sequence when $\delta_d$ = $6$ or $\delta_d$ = $10$, and therefore  the results of \cite{wang2021pole} shown in this table are obtained when $\delta_d$ = $1.5$.}
\Huge
\resizebox{2.0\columnwidth}{!}{
\begin{tabular}
{rr|ccccc|ccccc|ccccc|ccccc}
\toprule
    &
    & \multicolumn{5}{c|}{$\Delta_{\mathrm{pos}}$}
    & \multicolumn{5}{c|}{RMSE$_{\mathrm{pos}}$}
    & \multicolumn{5}{c|}{$\Delta_{\mathrm{ang}}$}
    & \multicolumn{5}{c}{RMSE$_{\mathrm{ang}}$} \\
    &
    & \multicolumn{5}{c|}{[${m}$]}
    & \multicolumn{5}{c|}{[${m}$]}
    & \multicolumn{5}{c|}{[${\degree}$]}
    & \multicolumn{5}{c}{[${\degree}$]} \\
\toprule
$\delta_{d}$&\text{Sequence}
& \text{ \citep{wang2021pole}} 
& \text{ \citep{dong2021online}} & \text{PF}
& \text{I-PF} & \text{I+N-PF}
& \text{ \citep{wang2021pole}} 
& \text{ \citep{dong2021online}} & \text{PF}
& \text{I-PF} & \text{I+N-PF}
& \text{ \citep{wang2021pole}} 
& \text{ \citep{dong2021online}} & \text{PF}
& \text{I-PF} & \text{I+N-PF}
& \text{ \citep{wang2021pole}} 
& \text{ \citep{dong2021online}} & \text{PF}
& \text{I-PF} & \text{I+N-PF}
\\ 
\bottomrule 
\multirow{11}{*}{\text{6}}
 & 11 & 4.074* & 3.727 & 1.201 & 1.188 & 1.136 & 7.509* & 5.239 & 1.524 & 1.537 & 1.427 & 1.199* & 1.387 & 0.609 & 0.693 & 0.580 & 1.892* & 2.082 & 1.050 & 1.112 &  0.987\\
 & 12 & 0.691* & 5.277 & 2.623 & 2.307 & 2.411 & 2.056* & 6.248 & 3.840 & 3.230 & 3.516 & 0.031* & 0.735 & 0.452 & 0.408 & 0.393 & 0.134* & 0.926 & 0.582 & 0.543 &  0.534\\
 & 13 & 7.132 & 1.665 & 0.785 & 0.723 & 0.655 & 12.255 & 2.967 & 1.124 & 0.983 & 0.843 & 2.072 & 1.058 & 0.561 & 0.547 & 0.459 & 3.400 & 2.011 & 1.129 & 1.056 &  0.921\\
 & 14 & 4.482* & 2.983 & 1.436 & 1.151 & 1.120 & 5.418* & 4.036 & 2.056 & 1.567 & 1.562 & 5.010* & 4.083 & 2.508 & 2.205 & 2.133 & 6.949* & 5.486 & 3.932 & 3.575 &  3.521\\
 & 15 & 15.662* & 2.629 & 1.705 & 1.162 & 1.227 & 22.409* & 3.818 & 2.299 & 1.497 & 1.589 & 4.652* & 2.559 & 2.035 & 1.523 & 1.704 & 6.384* & 3.760 & 3.093 & 2.359 &  2.577\\
 & 16 & 2.387 & 1.373 & 0.647 & 0.634 & 0.625 & 4.603 & 1.910 & 0.839 & 0.805 & 0.804 & 0.743 & 0.946 & 0.563 & 0.557 & 0.496 & 1.724 & 1.687 & 0.983 & 0.951 &  0.918\\
 & 17 & 8.728 & 1.866 & 0.801 & 0.763 & 0.743 & 10.504 & 3.123 & 1.148 & 1.040 & 1.014 & 0.381 & 0.199 & 0.221 & 0.227 & 0.225 & 0.475 & 0.255 & 0.278 & 0.287 &  0.273\\
 & 18 & 1.582* & 4.107 & 1.323 & 1.135 & 1.159 & 3.216* & 5.635 & 1.766 & 1.512 & 1.531 & 0.670* & 1.210 & 0.643 & 0.581 & 0.563 & 1.241* & 1.509 & 0.886 & 0.788 &  0.729\\
 & 19 & 19.791 & 4.537 & 1.722 & 1.470 & 1.446 & 33.487 & 8.315 & 3.474 & 3.003 & 3.091 & 5.249 & 1.928 & 0.947 & 0.862 & 0.847 & 8.610 & 3.587 & 1.749 & 1.597 &  1.631\\
 & 20 & 1.629* & 3.343 & 1.230 & 1.059 & 1.036 & 2.897* & 4.084 & 1.673 & 1.355 & 1.402 & 0.263* & 0.587 & 0.298 & 0.332 & 0.301 & 0.379* & 0.748 & 0.408 & 0.451 &  0.410\\
 & 21 & 19.583* & 9.711 & 3.106 & 2.266 & 2.183 & 26.579* & 14.192 & 4.580 & 3.229 & 3.091 & 0.402* & 0.732 & 0.312 & 0.333 & 0.345 & 0.598* & 1.659 & 0.602 & 0.637 &  0.635\\
\bottomrule 
 & Average  & 7.795 & 3.747 & 1.507 & 1.260 & 1.249 & 11.903 & 5.415 & 2.211 & 1.796 & 1.806 & 1.879 & 1.402 & 0.832 & 0.752 & 0.731 & 2.890 & 2.155 & 1.336 & 1.214 &  1.194\\
\bottomrule\
\multirow{11}{*}{\text{10}}
 & 11 & 4.074* & 5.058 & 1.503 & 1.343 & 1.250 & 7.509* & 6.440 & 1.987 & 1.673 & 1.565 & 1.199* & 2.001 & 0.707 & 0.704 & 0.593 & 1.892* & 2.960 & 1.243 & 1.167 &  1.080\\
 & 12 & 0.691* & 15.431 & 3.045 & 2.836 & 3.224 & 2.056* & 20.229 & 3.903 & 3.883 & 4.432 & 0.031* & 1.776 & 0.617 & 0.530 & 0.552 & 0.134* & 2.052 & 0.775 & 0.700 &  0.723\\
 & 13 & 9.276 & 2.792 & 0.970 & 0.903 & 0.849 & 14.390 & 4.486 & 1.315 & 1.153 & 1.100 & 2.367 & 1.347 & 0.673 & 0.652 & 0.626 & 3.641 & 2.285 & 1.288 & 1.204 &  1.196\\
 & 14 & 4.482* & 3.462 & 1.933 & 1.529 & 1.520 & 5.418* & 4.117 & 2.291 & 1.878 & 1.880 & 5.010* & 4.077 & 2.705 & 2.615 & 2.751 & 6.949* & 5.285 & 3.623 & 3.736 &  3.704\\
 & 15 & 15.662* & 9.070 & 3.188 & 1.967 & 1.855 & 22.409* & 11.129 & 4.295 & 2.491 & 2.498 & 4.652* & 4.328 & 2.592 & 2.093 & 2.134 & 6.384* & 5.626 & 3.578 & 2.936 &  3.168\\
 & 16 & 1.478* & 2.490 & 1.063 & 1.017 & 0.919 & 3.175* & 3.793 & 1.421 & 1.348 & 1.235 & 0.118* & 1.578 & 0.932 & 0.969 & 0.855 & 0.313* & 2.788 & 1.566 & 1.645 &  1.464\\
 & 17 & 8.855 & 4.622 & 1.252 & 1.086 & 1.118 & 10.652 & 6.689 & 1.797 & 1.441 & 1.504 & 0.352 & 0.284 & 0.256 & 0.254 & 0.232 & 0.442 & 0.345 & 0.313 & 0.321 &  0.305\\
 & 18 & 1.582* & 4.317 & 2.860 & 1.777 & 1.698 & 3.216* & 5.426 & 4.002 & 2.431 & 2.229 & 0.670* & 1.076 & 0.986 & 0.850 & 0.815 & 1.241* & 1.344 & 1.235 & 1.074 &  1.033\\
 & 19 & 1.758* & 11.250 & 2.680 & 2.097 & 1.782 & 4.324* & 17.368 & 4.986 & 3.706 & 2.947 & 0.535* & 3.072 & 1.289 & 1.114 & 1.140 & 1.785* & 6.180 & 2.183 & 1.812 &  1.853\\
 & 20 & 1.629* & 4.316 & 1.392 & 1.259 & 1.165 & 2.897* & 5.174 & 1.787 & 1.564 & 1.423 & 0.263* & 0.413 & 0.454 & 0.488 & 0.434 & 0.379* & 0.502 & 0.608 & 0.696 &  0.607\\
 & 21 & 19.583* & 12.991 & 4.473 & 3.470 & 3.020 & 26.579* & 22.479 & 6.591 & 5.205 & 4.533 & 0.402* & 1.101 & 0.365 & 0.348 & 0.359 & 0.598* & 2.200 & 0.692 & 0.670 &  0.638\\
\bottomrule 
 &  Average  & 6.279 & 6.891 & 2.214 & 1.753 & 1.673 & 9.330 & 9.757 & 3.125 & 2.434 & 2.304 & 1.418 & 1.914 & 1.052 & 0.965 & 0.954 & 2.160 & 2.870 & 1.555 & 1.451 &  1.434\\
\bottomrule 
\end{tabular}
}
\label{pole_poseang}
\end{table*}

\begin{figure*}[t]
  \centering
  \includegraphics[width=0.99\linewidth]{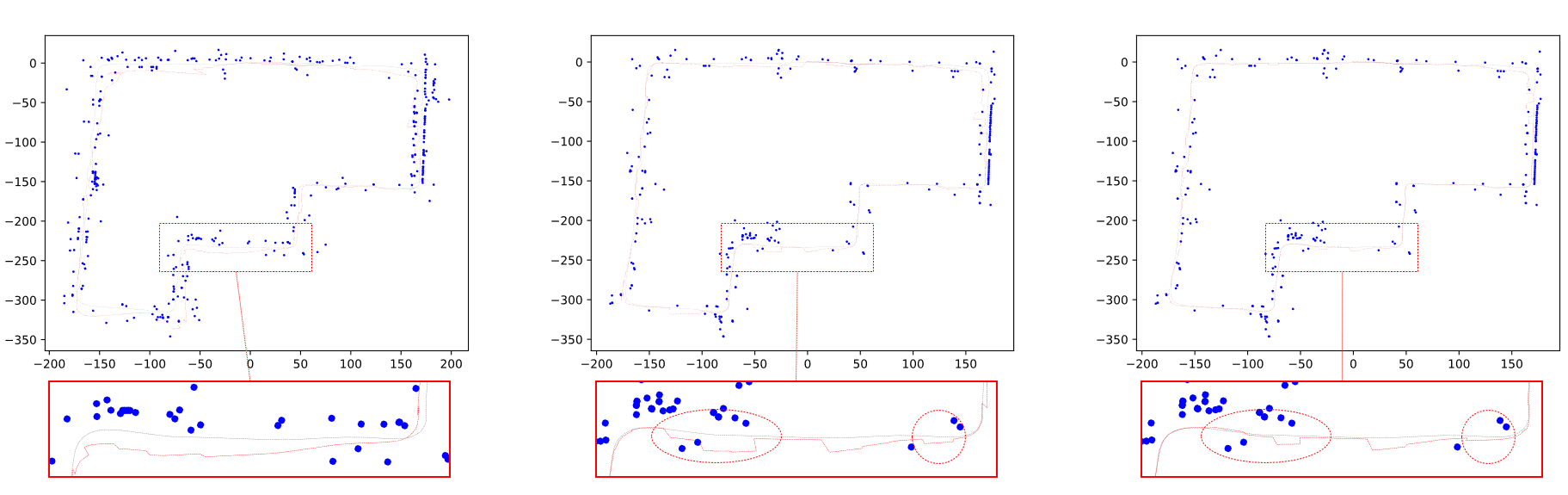}
  \caption{The visualization of localization trajectory of \cite{dong2021online} and our methods in SemanticKITTI \citep{behley2019semantickitti} sequence $15$. The ground truth trajectory is shown in grey and the estimated trajectory is shown in red. Left: The localization trajectory with baseline map \citep{dong2021online} and normal particle filter \citep{dellaert1999monte}. Middle: The localization trajectory with our pole-map and normal particle filter. Right: The localization trajectory with our pole-map and our semantic-aware particle filter with both semantic inconsistency and nearest neighbor schemes. In this experiment, $\phi_{odo}$ is set to $40$\% and $\phi_{O}$ is set to $80$\%.}
  \label{fig:localization}
\end{figure*}

\begin{figure}[t]
  \centering
  \includegraphics[width=0.99\columnwidth]{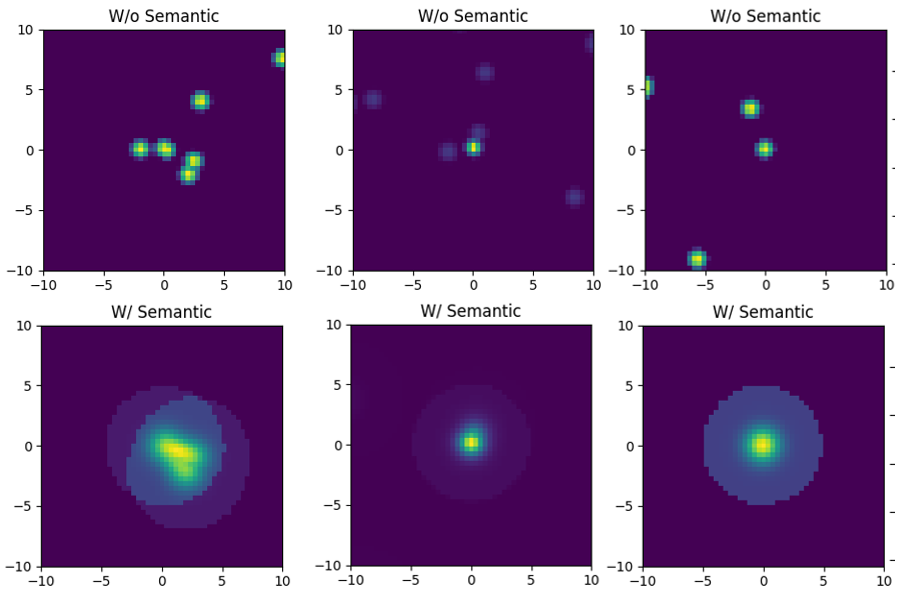}
  \caption{The visualization of particle weights in particle filter estimated with and without semantic information. The particles are distributed in the square of its vicinity. The lighter color represents the higher weights. Top: The weights of normal particle filter \citep{dellaert1999monte}. Bottom: The weights of our semantic-aware particle filter with both semantic inconsistency and nearest neighbor schemes.}
  \label{fig:particle_weights}
\end{figure}

\textbf{Results.}
Table \ref{pole_poseang} shows that our pole-map is better in terms of localization than the baselines. 
\cite{dong2021online} utilize LiDAR scans to extract pole-like landmarks. Comparing this baseline method with our \textbf{PF}, in all the sequences, localization in our pole-maps achieves better performance. 
This means that compared with the baseline method which only exploits vertical structures in the pole-map without considering their actual semantics, our pole-map extracts landmarks useful for localization from our semantic segmentation. 
As shown in  Table \ref{pole_number} from Section \nameref{exp:mapping}, the baseline method provides fewer poles in each frame but generates a similar number of landmarks in the final pole-map, which means that it obtains less consistent pole-like objects.
The landmarks we extracted are more consistent across different frames than the baseline method. 

\cite{wang2021pole} propose utilizing the LiDAR semantics predicted by a semantic segmentation model for mapping and localization. Empirically, we find that this method must rely on enough pole-like landmarks to obtain semantic descriptors and match them in localization, otherwise it is easy to fail. 
Because the larger $\delta_d$ results in fewer pole-like landmarks in the map, in some of the sequences where there are not enough poles it fails to find the matched landmarks across the throughout sequences, and the localization never succeed. In these cases, we set $\delta_d$ to $1.5$ for denser poles and mark the localization results in this setting with $*$. Additionally, we set $\phi_{O}$ as $0$\% in its experiments.  

Comparing this baseline method with our \textbf{I+N-PF}, we achieve better performance with lower position $\Delta$ in most sequences based on our pole-maps. 
In the sequence $12$ where our method is worse, the number of frames used for mapping in \cite{wang2021pole} are $4$ to $6$ times ours. 
In the other sequences, although with more frames for mapping and lower observation uncertainty $\phi_{O}$, \cite{wang2021pole} cannot achieve higher performance. The results show that our method is better than the ICP-based localization which requires confident matching.

As shown in Fig. \ref{fig:localization}, the baseline method \citep{dong2021online} drifts along the route (left), as the inconsistent poles provide inconsistent updating for particle weights, which hinders the particles convergence. On contrary, our pole segmentation provides consistent poles along the route and makes particle convergent to the ground truth location (middle, right). However, as the odometry is really noisy and the observation makes false negative detection simulated by us, the particles with dispersed distribution estimate inaccurate pose (middle). When incorporating semantic information into particle filter, the particles with more discriminative weight are easier to concentrated to the ground truth poses as we expected (right). As shown in Fig. \ref{fig:particle_weights}, while the normal particle filter without semantic information has activation at multiple centers, our
semantic-aware particle filter with both semantic inconsistency
and nearest neighbor schemes provides more deterministic estimation.
Quantitatively, 
in Table \ref{pole_poseang}, 
we can see that the semantic particle filter improves the localization performance compared with the one without semantics. For example, when $\delta_d$ is set to $10$, the semantic-aware inconsistency \textbf{I-PF} reduces the position $\Delta$ by 0.462 ($20.84\%$) and heading $\Delta$ by 0.087 ($8.28\%$). When incorporating both the semantic-aware inconsistency and semantic-aware nearest neighbor \textbf{I+N-PF}, the semantic-aware particle filter reduces the position $\Delta$ by 0.542 ($24.46\%$) and heading $\Delta$ by 0.099 ($9.38\%$).
Noted that 
in this experimental setting, the odometry noise level is $\phi_{odo}$ is set to $40$\% and the online landmark observation noise level $\phi_{O}$ are set to $80$\%. 
In ablation studies, we will have a detailed analysis of the different settings of mapping, different noises in localization, and different ways to utilize semantic information.

\section{Ablation Study}
In this section, we investigate (i) the mapping and localization performance with and without multi-layer mapping, (ii) the localization performance under different choices of $\delta_d$ (the distance between two keyframes during mapping), (iii) the localization performance under different odometry noise $\phi_{odo}$ and observation noise $\phi_{O}$, and (iv) how our pole-map and semantic information benefit the localization performance. 

\subsection{Multi-Layer Mapping}\label{sec:multilayermapping}
In our proposed methods, semantic information is utilized to improve particle-filter localization. However, it is often inevitable that semantics can contain noise from segmentation. When incorporating imperfect semantic information into localization, the noise from pole segmentation can degrade the localization performance. 
To deal with this issue, we propose multi-layer mapping to reduce the semantic ambiguity in the pole-map and improve the semantic-aware localization performance with this map. To investigate the effectiveness, we compare the mapping and localization performance with normal single-layer mapping and the proposed multi-layer mapping. In this experiment $\phi_{odo}$ is set to $40$\%. Because this experiment is designed only for evaluating mapping performance, $\phi_{O}$ are set to $0$\% without simulating online pole detection noise. Table \ref{tbl:multilayermapping} shows that multi-layer mapping improves both precision and recall for mapping, indicating it provides more TPs without introducing many FPs and FNs. More importantly, as shown in Table \ref{tbl:multilayerlocalization}, localization performance with multi-layer mapping is improved when incorporating semantic information into the particle-filter localization, indicating the robustness of multi-layer mapping while the semantic information is noisy.

\begin{table}[t]
\caption[Multi Layer Mapping]{The comparison of mapping performance in SemanticKITTI \citep{behley2019semantickitti} sequence 01 and 08 by our methods using single-layer and multi-layer mapping. $\textbf{S.}$ denotes the single-layer mapping as in \cite{dong2021online}. $\textbf{M.}$ denotes our proposed multi-layer mapping.}
\centering
\resizebox{0.58\columnwidth}{!}{
\begin{tabular}
{|l|cc|cc|}
\toprule
Sequence
& \multicolumn{2}{c|}{01}
& \multicolumn{2}{c|}{08}\\
\midrule
Method
& \textbf{S.}
& \textbf{M.}
& \textbf{S.}
& \textbf{M.}\\
\midrule
Precision
& 0.68 & 0.73 & 0.74 & 0.76
\\
\midrule
Recall
& 0.39 & 0.56 & 0.75 & 0.86
\\
\midrule
F1
& 0.49 & 0.63 & 0.74 & 0.81
\\
\bottomrule
\end{tabular}
}
\label{tbl:multilayermapping}
\end{table}


\begin{table}[t]
\caption[Multi Layer Mapping]{The comparison of localization performance in SemanticKITTI \citep{behley2019semantickitti} sequence 11-21 by our methods using single layer and multi-layer mapping. $\textbf{S.}$ denotes the single layer mapping as in \cite{dong2021online}. $\textbf{M.}$ denotes our proposed multi-layer mapping. In this experiment $\phi_{odo}$ is set to $40$\% and $\phi_{O}$ are set to $0$\% without simulated pole extraction noise.}
\centering
\resizebox{0.78\columnwidth}{!}{
\begin{tabular}
{|l|cccc|cccc|}
\toprule
Sequence
& \multicolumn{4}{c|}{11-21}
\\
\midrule
Method
& PF/\textbf{S.}
& PF/\textbf{M.}
& I+N-PF/\textbf{S.}
& I+N-PF/\textbf{M.}
\\
\midrule
$\Delta_{\mathrm{pos}}$
 & 0.899 & 0.914 & 0.884 & 0.869 
\\
\midrule
RMSE$_{\mathrm{pos}}$ 
 & 1.302 & 1.363 & 1.246 & 1.265 
\\
\midrule
$\Delta_{\mathrm{ang}}$
 & 0.551 & 0.528 & 0.534 & 0.522 
\\
\midrule
RMSE$_{\mathrm{ang}}$ 
 & 0.980 & 0.958 & 0.948 & 0.979 
\\
\bottomrule
\end{tabular}
}
\label{tbl:multilayerlocalization}
\end{table}

\subsection{The Effect of $\delta_d$}
\label{sec:mappingdist}
\begin{figure}[t]
  \centering
  \includegraphics[width=0.99\linewidth]{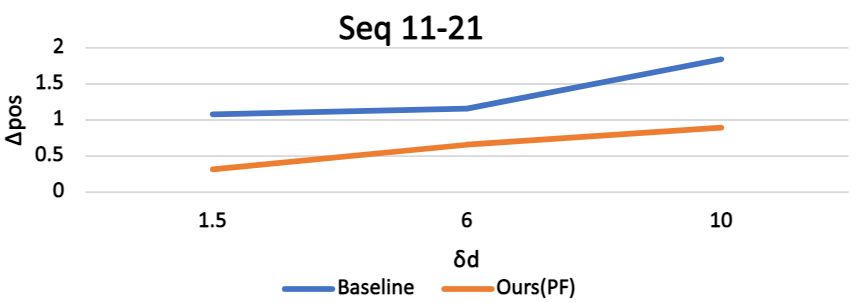}
  \caption{The comparison of localization performance when choosing different distance threshold $\delta_d$. The baseline method \citep{dong2021online} (blue) and our method without semantics (orange) are compared. In the experiments $\phi_{odo}$ is set to $40$\% and $\phi_{O}$ are set to $0$\% without simulated pole extraction noise.
  }
  \label{fig:mapping_distance}
\end{figure}

As mentioned before, the distance $\delta_d$ is designed to split a SemanticKITTI sequence into non-overlapping subsets as mapping and localization data, respectively, to evaluate the localization performance. To investigate a reasonable $\delta_d$ we compare the localization performance of baseline \citep{dong2021online} and our method with different $\delta_d$. In the experiments $\phi_{odo}$ is set to $40$\% and $\phi_{O}$ are set to $0$\%. Our method is based on a normal particle filter localization for comparison. As shown in Fig. \ref{fig:mapping_distance}, the mean absolute error ($\Delta$) increases when $\delta_d$ increases.
Throughout the paper, $\delta_d$ is set to $10$ without specification.

\begin{table*}[t]
\caption[Error Condition]{The comparison of localization performance of normal  particle-filter \citep{dellaert1999monte} and the proposed semantic-aware particle-filter. Localization performance is evaluated at different levels of odometry noise $\phi_{odo}$ and observation noise $\phi_{O}$. The same as in Table \ref{pole_poseang}, \textbf{PF} denotes the normal particle filter localization \citep{dellaert1999monte}. \textbf{I-PF} denotes semantic particle filter localization by incorporating the Semantic-Aware Inconsistency. \textbf{I+N-PF} denotes semantic particle filter localization by incorporating the Semantic-Aware Inconsistency and the Semantic-Aware Nearest Neighbor. \textbf{Ipr.} denotes the improvement of I+N-PF compared with PF.}
\Huge
\resizebox{2.1\columnwidth}{!}{
\begin{tabular}
{c|cc|cccc|cccc|cccc|cccc}
\toprule
&&
    & \multicolumn{4}{c|}{$\Delta_{\mathrm{pos}}$}
    & \multicolumn{4}{c|}{RMSE$_{\mathrm{pos}}$}
    & \multicolumn{4}{c|}{$\Delta_{\mathrm{ang}}$}
    & \multicolumn{4}{c}{RMSE$_{\mathrm{ang}}$} \\
&&
    & \multicolumn{4}{c|}{[{m}]}
    & \multicolumn{4}{c|}{[{m}]}
    & \multicolumn{4}{c|}{[{$\degree$}]}
    & \multicolumn{4}{c}{[{$\degree$}]} \\
\toprule
\text{$\delta_{d}$}
& \text{$\phi_{odo}$}
& \text{$\phi_{O}$}
& \text{PF}
& \text{I-PF} & \text{I+N-PF}
& \text{Ipr.}
& \text{PF}
& \text{I-PF} & \text{I+N-PF}
& \text{Ipr.}
& \text{PF}
& \text{I-PF} & \text{I+N-PF}
& \text{Ipr.}
& \text{PF}
& \text{I-PF} & \text{I+N-PF}
& \text{Ipr.}
\\
\bottomrule
 \multirow{8}{*}{\text{6}}
 & \multirow{4}{*}{\text{0.2}}
 & \text{0}
 & 0.381 & 0.380 & 0.373 & 2.25\% & 0.515 & 0.510 & 0.495 & 3.85\% & 0.250 & 0.245 & 0.245 & 1.87\% & 0.405 & 0.403 & 0.402 & 0.87\%\\
 & & \text{0.2} 
 & 0.404 & 0.396 & 0.397 & 1.96\% & 0.537 & 0.523 & 0.528 & 1.71\% & 0.264 & 0.249 & 0.257 & 2.74\% & 0.423 & 0.405 & 0.414 & 1.95\%\\
 & & \text{0.5}
 & 0.443 & 0.440 & 0.438 & 1.12\% & 0.591 & 0.581 & 0.577 & 2.47\% & 0.280 & 0.283 & 0.275 & 1.73\% & 0.441 & 0.445 & 0.433 & 1.94\%\\
 & & \text{0.8}
 & 0.651 & 0.613 & 0.601 & 7.72\% & 0.926 & 0.846 & 0.828 & 10.63\% & 0.411 & 0.394 & 0.398 & 3.09\% & 0.633 & 0.611 & 0.619 & 2.19\%\\
\cmidrule{2-19}
 & \multirow{4}{*}{\text{0.4}}
 & \text{0}
 & 0.675 & 0.660 & 0.651 & 3.56\% & 0.972 & 0.930 & 0.921 & 5.28\% & 0.404 & 0.408 & 0.403 & 0.10\% & 0.718 & 0.718 & 0.716 & 0.28\%\\
 & & \text{0.2}
 & 0.731 & 0.701 & 0.703 & 3.72\% & 1.048 & 0.988 & 1.005 & 4.06\% & 0.474 & 0.447 & 0.457 & 3.46\% & 0.832 & 0.789 & 0.821 & 1.22\%\\
 & & \text{0.5}
 & 0.888 & 0.837 & 0.830 & 6.47\% & 1.357 & 1.246 & 1.236 & 8.93\% & 0.546 & 0.522 & 0.531 & 2.66\% & 0.957 & 0.916 & 0.923 & 3.54\%\\
 & & \text{0.8}
 & 1.507 & 1.260 & 1.249 & 17.11\% & 2.211 & 1.796 & 1.806 & 18.30\% & 0.832 & 0.751 & 0.732 & 12.06\% & 1.336 & 1.214 & 1.194 & 10.58\%\\
\bottomrule
 \multirow{8}{*}{\text{10}}
 & \multirow{4}{*}{\text{0.2}}
 & \text{0}
 & 0.483 & 0.482 & 0.478 & 0.90\% & 0.647 & 0.659 & 0.652 & -0.69\% & 0.301 & 0.302 & 0.285 & 5.19\% & 0.489 & 0.503 & 0.481 & 1.53\%\\
 & & \text{0.2}
 & 0.510 & 0.506 & 0.499 & 2.05\% & 0.698 & 0.684 & 0.680 & 2.56\% & 0.323 & 0.313 & 0.311 & 3.67\% & 0.530 & 0.514 & 0.517 & 2.43\%\\
 & & \text{0.5}
 & 0.598 & 0.594 & 0.576 & 3.70\% & 0.810 & 0.785 & 0.762 & 5.87\% & 0.362 & 0.360 & 0.345 & 4.93\% & 0.573 & 0.566 & 0.548 & 4.37\%\\
 & & \text{0.8}
 & 0.957 & 0.857 & 0.827 & 13.62\% & 1.308 & 1.186 & 1.140 & 12.87\% & 0.502 & 0.484 & 0.475 & 5.47\% & 0.745 & 0.729 & 0.713 & 4.35\%\\
\cmidrule{2-19}
 & \multirow{4}{*}{\text{0.4}}
 & \text{0}
 & 0.914 & 0.874 & 0.869 & 4.88\% & 1.363 & 1.264 & 1.265 & 7.23\% & 0.528 & 0.527 & 0.522 & 1.13\% & 0.958 & 0.967 & 0.979 & -2.29\%\\
 & & \text{0.2}
 & 1.011 & 0.957 & 0.933 & 7.70\% & 1.579 & 1.481 & 1.410 & 10.65\% & 0.574 & 0.560 & 0.558 & 2.73\% & 1.066 & 1.041 & 1.030 & 3.36\%\\
 & & \text{0.5}
 & 1.186 & 1.092 & 1.083 & 8.65\% & 1.698 & 1.533 & 1.508 & 11.17\% & 0.669 & 0.643 & 0.634 & 5.23\% & 1.119 & 1.088 & 1.070 & 4.41\%\\
 & & \text{0.8}
 & 2.215 & 1.753 & 1.673 & 24.46\% & 3.125 & 2.434 & 2.304 & 26.27\% & 1.052 & 0.965 & 0.954 & 9.38\% & 1.555 & 1.451 & 1.434 & 7.80\%\\
\bottomrule
\end{tabular}
}
\label{pole_posenoise}
\end{table*}

\subsection{Different Levels of Uncertainties $\phi_{odo}$ and $\phi_{O}$}\label{sec:expansion}
As mentioned in Section \nameref{sec:localization}, we add noise to the odometry and observation to simulate the noisy scenarios. In this section, we investigate how different noise levels influence localization performance.
When the poles are sparse or absent in a specific scene and the odometry used for particle pose prediction is subject to a large uncertainty at the same time, the estimated position can drift. If the drift is large enough, the particle weights can stop updating, as all the particles fail to find the corresponding poles in the data association. 
The noisier the odometry and online observation, the more frequently this situation happens. We chose different $\phi_{odo}$ and $\phi_{O}$ to investigate the effectiveness of semantic information in particle-filter localization in such scenarios.


As shown in Table \ref{pole_posenoise}, the localization performance drops with the increasing odometry noise $\phi_{odo}$ or observation noise $\phi_{O}$. When incorporating semantic information, the localization performance can be improved in almost all the choices of $\phi_{odo}$ and $\phi_{O}$. More importantly, semantic information can bring a larger gain in localization performance when confronted with larger uncertainties. 
This can be explained that semantic information can help the particles converge more quickly, as analyzed in Section \nameref{semantic_ic} and \nameref{semantic_nn}.

\subsection{Semantic-Aware Particle Filter Localization}
\label{sec:inconsistency}
Generally, localization is accurate and fast when the particles are densely concentrated with a single-mode distribution, which means most of the particles share the same set of associated landmarks in the map. Imagine that we will have only one set of correspondence between the online observation and map landmarks if all particles are very close to the ground-truth state. 

In practice, three additional quantitative metrics can be used to reflect the particle localization performance.  The first one is $N_{A^k}$, representing the number of sets of associations (number of different sets of $A^k$ in Eq. \ref{pd_corr}) for all $k$ particles. 
The second one is $\phi_{\bar{A}}$ representing the ratio of the number of correct landmark associations to all associations, which is empirically evaluated as the average of the similarity of an approximated correspondence being a correct one, i.e., through the $cos()$ term in Eq. \ref{particle_semantic_inconsitency_1}.
The third one $\phi_{\bar{y}}$ represents the ratio of the number of correspondences with consistent pole categories to the total number of correspondences.

We quantitatively investigate the localization performance along with the above three metrics above in the same experimental setting as in Section \nameref{sec:localization}. Finally, we take the average $N_{A^k}$, $\phi_{\bar{A}}$, and $\phi_{\bar{y}}$ of all frames and show the results in Table \ref{pole_correpondence}.
 Comparing particle filter with semantic-aware inconsistency (\textbf{I-PF}) with the one without semantic (\textbf{PF}), the number of $N_{A^k}$ is largely reduced. This shows that particles quickly converge when incorporating semantic-aware inconsistency in Eq. \ref{particle_measurement_counter} as we expected. Meanwhile, the association accuracy $\phi_{\bar{A}}$ and category accuracy $\phi_{\bar{y}}$ are improved, indicating that the particles find more accurate correspondences, and thus higher localization performance is achieved. 
Interestingly, compared with \textbf{I-PF}, although the particle filter with semantic-aware nearest neighbor (\textbf{N-PF}) achieve larger reduction in $N_{A^k}$ and larger improvement in $\phi_{\bar{y}}$, it cannot achieve larger improvement at $\phi_{\bar{A}}$ and localization performance $\Delta_{\mathrm{pos}}$. This shows that the semantic nearest neighbor working alone brings limited performance improvement because the semantic nearest neighbor may mislead when the observation is incorrect. However it works well with semantic-aware inconsistency \textbf{I+N-PF} because the erroneous correspondences in such situation can be suppressed, which takes both advantages and achieves the most improvement.


\begin{table*}[t]
\caption[Pole-map creation]{The quantitative results of the proposed semantic-aware  particle-filter localization. Results are evaluated with respect to $N_{A^k}$, $\phi_{\bar{A}}$, and $\phi_{\bar{y}}$ from Section \nameref{sec:inconsistency}. Localization performance $\Delta_{pos}$ in Section \nameref{sec:localization} is also compared. The same as in Table \ref{pole_poseang}, \textbf{PF} denotes the normal particle filter localization \citep{dellaert1999monte}. \textbf{I-PF} denotes semantic particle filter localization by incorporating the Semantic-Aware Inconsistency. \textbf{N-PF} denotes semantic particle filter localization by incorporating the Semantic-Aware Nearest Neighbor. \textbf{I+N-PF} denotes semantic particle filter localization by incorporating the Semantic-Aware Inconsistency and the Semantic-Aware Nearest Neighbor. In this experiment, $\phi_{odo}$ is set to $40$\% and $\phi_{O}$ is set to $80$\%.}.
\resizebox{2.1\columnwidth}{!}{
\begin{tabular}
{rr|cccc|cccc|cccc|cccc}
\toprule
&
    & \multicolumn{4}{c|}{$\Delta_{\mathrm{pos}}$}
    & \multicolumn{4}{c|}{$N_{A^k}$}
    & \multicolumn{4}{c|}{$\phi_{\bar{A}}$}
    & \multicolumn{4}{c}{$\phi_{\bar{y}}$}\\
\toprule
\text{$\delta_d$}
& \text{Sequence}
& \text{PF} 
& \text{I-PF} & \text{N-PF} & \text{I+N-PF}
& \text{PF} 
& \text{I-PF} & \text{N-PF} & \text{I+N-PF}
& \text{PF} 
& \text{I-PF} & \text{N-PF} & \text{I+N-PF}
& \text{PF} 
& \text{I-PF} & \text{N-PF} & \text{I+N-PF}
\\ 
\bottomrule
 \multirow{11}{*}{\text{6}}
 & 11 & 1.201 & 1.188 & 1.213 & 1.136 & 14.592 & 12.317 & 6.845 & 6.014 & 0.856 & 0.878 & 0.881 & 0.903 & 0.759 & 0.785 & 0.906 & 0.930\\
 & 12 & 2.623 & 2.307 & 2.742 & 2.411 & 6.712 & 5.513 & 5.098 & 4.219 & 0.816 & 0.850 & 0.819 & 0.854 & 0.786 & 0.819 & 0.831 & 0.867\\
 & 13 & 0.785 & 0.723 & 0.721 & 0.655 & 38.009 & 35.991 & 16.400 & 15.714 & 0.879 & 0.886 & 0.916 & 0.924 & 0.746 & 0.753 & 0.956 & 0.964\\
 & 14 & 1.436 & 1.151 & 1.227 & 1.120 & 19.679 & 12.500 & 15.018 & 10.375 & 0.875 & 0.908 & 0.877 & 0.910 & 0.867 & 0.901 & 0.890 & 0.923\\
 & 15 & 1.705 & 1.162 & 1.414 & 1.227 & 12.580 & 9.689 & 6.909 & 5.420 & 0.798 & 0.839 & 0.795 & 0.843 & 0.732 & 0.774 & 0.816 & 0.865\\
 & 16 & 0.647 & 0.634 & 0.630 & 0.625 & 12.985 & 12.179 & 7.129 & 6.738 & 0.906 & 0.913 & 0.930 & 0.936 & 0.842 & 0.849 & 0.960 & 0.966\\
 & 17 & 0.801 & 0.763 & 0.852 & 0.743 & 34.248 & 33.745 & 18.303 & 17.903 & 0.917 & 0.917 & 0.936 & 0.939 & 0.798 & 0.798 & 0.965 & 0.968\\
 & 18 & 1.323 & 1.135 & 1.349 & 1.159 & 16.955 & 15.784 & 8.640 & 7.946 & 0.862 & 0.877 & 0.868 & 0.889 & 0.769 & 0.783 & 0.907 & 0.929\\
 & 19 & 1.722 & 1.470 & 1.967 & 1.446 & 14.714 & 12.911 & 8.101 & 7.132 & 0.871 & 0.886 & 0.878 & 0.902 & 0.819 & 0.834 & 0.905 & 0.930\\
 & 20 & 1.230 & 1.059 & 1.150 & 1.036 & 6.500 & 5.853 & 4.345 & 3.897 & 0.830 & 0.856 & 0.840 & 0.874 & 0.774 & 0.798 & 0.864 & 0.899\\
 & 21 & 3.106 & 2.266 & 2.830 & 2.183 & 4.730 & 4.134 & 2.988 & 2.692 & 0.827 & 0.860 & 0.827 & 0.864 & 0.781 & 0.809 & 0.840 & 0.878\\
\bottomrule
 & Average & 1.507 & 1.260 & 1.463 & 1.249 & 16.518 & 14.601 & 9.070 & 8.004 & 0.858 & 0.879 & 0.870 & 0.894 & 0.788 & 0.809 & 0.894 & 0.920\\
\bottomrule
 \multirow{11}{*}{\text{10}}
 & 11 & 1.503 & 1.343 & 1.427 & 1.250 & 17.824 & 14.929 & 8.788 & 7.176 & 0.763 & 0.801 & 0.763 & 0.824 & 0.674 & 0.709 & 0.797 & 0.861\\
 & 12 & 3.045 & 2.836 & 3.798 & 3.224 & 9.216 & 6.577 & 6.538 & 5.250 & 0.761 & 0.814 & 0.749 & 0.806 & 0.735 & 0.788 & 0.758 & 0.817\\
 & 13 & 0.970 & 0.903 & 0.915 & 0.849 & 57.153 & 52.843 & 27.808 & 25.815 & 0.834 & 0.846 & 0.865 & 0.878 & 0.699 & 0.710 & 0.905 & 0.919\\
 & 14 & 1.933 & 1.529 & 2.519 & 1.520 & 24.091 & 19.121 & 18.121 & 14.970 & 0.750 & 0.795 & 0.735 & 0.796 & 0.755 & 0.802 & 0.751 & 0.813\\
 & 15 & 3.188 & 1.967 & 3.298 & 1.855 & 14.916 & 11.252 & 8.443 & 6.435 & 0.623 & 0.724 & 0.607 & 0.728 & 0.586 & 0.681 & 0.629 & 0.755\\
 & 16 & 1.063 & 1.017 & 1.001 & 0.919 & 20.690 & 18.095 & 10.696 & 9.975 & 0.819 & 0.844 & 0.832 & 0.859 & 0.745 & 0.771 & 0.859 & 0.887\\
 & 17 & 1.252 & 1.086 & 1.307 & 1.118 & 39.057 & 36.598 & 23.402 & 22.333 & 0.899 & 0.904 & 0.907 & 0.912 & 0.804 & 0.809 & 0.944 & 0.949\\
 & 18 & 2.860 & 1.777 & 2.344 & 1.698 & 30.481 & 24.579 & 15.917 & 12.571 & 0.776 & 0.806 & 0.781 & 0.829 & 0.686 & 0.712 & 0.824 & 0.876\\
 & 19 & 2.680 & 2.097 & 3.086 & 1.782 & 19.977 & 16.182 & 11.594 & 9.654 & 0.807 & 0.842 & 0.809 & 0.851 & 0.756 & 0.791 & 0.833 & 0.876\\
 & 20 & 1.392 & 1.259 & 1.312 & 1.165 & 8.829 & 7.257 & 5.743 & 4.557 & 0.743 & 0.786 & 0.756 & 0.806 & 0.673 & 0.706 & 0.786 & 0.839\\
 & 21 & 4.473 & 3.470 & 5.480 & 3.020 & 6.014 & 4.611 & 3.978 & 3.024 & 0.775 & 0.827 & 0.760 & 0.828 & 0.741 & 0.785 & 0.770 & 0.841\\
\bottomrule 
 & Average & 2.215 & 1.753 & 2.408 & 1.673 & 22.568 & 19.277 & 12.821 & 11.069 & 0.777 & 0.817 & 0.778 & 0.829 & 0.714 & 0.751 & 0.805 & 0.858\\
\bottomrule
\end{tabular}
}
\label{pole_correpondence}
\end{table*}




\section{Conclusion}  
\label{sec:conclusion}
In this work, we propose a full framework for semantic mapping and localization where the localization is achieved based on semantic particle filtering in
a multi-layer semantic pole-map created offline by a multi-channel LiDAR sensor. The semantic pole-map is built based on the pole semantics extracted by an efficient semantic segmentation method in the mask-classification paradigm. On applying this semantic pole-map for online localization, we have proposed a semantic particle-filter based scheme with poles as observations. We have both theoretically and empirically shown that our semantic particle-filter localization method given the semantic pole-map achieves very promising performance even with significant levels of uncertainties. 
In the future, we will investigate utilizing other semantic categories in our pole-map to improve localization performance.

{
\bibliographystyle{SageH}
\bibliography{egbib}
}

\end{document}